\documentclass[letterpaper]{article} 
\usepackage[preprint]{aaai2027}  
\usepackage[hyphens]{url}  
\usepackage{graphicx} 
\usepackage{natbib}  
\usepackage{caption} 
\usepackage{algorithm}
\usepackage{algorithmic}

\usepackage{newfloat}
\usepackage{listings}
\DeclareCaptionStyle{ruled}{labelfont=normalfont,labelsep=colon,strut=off} 
\floatstyle{ruled}
\newfloat{listing}{tb}{lst}{}
\floatname{listing}{Listing}

\usepackage{booktabs}

\usepackage[utf8]{inputenc}
\usepackage[T1]{fontenc}
\usepackage{amsmath,amssymb,amsfonts}
\usepackage{booktabs}
\usepackage{graphicx}
\usepackage{xcolor}
\newcommand{\funccode}{\textsc{FuncCode}}

\title{FuncCode: Compressing Kolmogorov--Arnold Networks in\\
Function Space with Hardware-Aware Quantization}
\author{
    Kazi Ahmed Asif Fuad\textsuperscript{\rm 1},
    Lizhong Chen\textsuperscript{\rm 1}
}

\affiliations{
    \textsuperscript{\rm 1}Department of EECS\\
    Oregon State University\\
    Corvallis, OR 97331\\
    fuadk@oregonstate.edu, chenliz@oregonstate.edu
}

\begin{document}

\maketitle

\begin{abstract}
Kolmogorov--Arnold Networks (KANs) replace scalar edge weights with learnable
univariate functions, increasing flexibility but also parameter memory because
each edge stores multiple coefficients, often together with a separate base
branch. We introduce FuncCode, a basis-agnostic compression approach that
forms shared codebooks from sampled edge responses, codes the basis and base
branches independently, and exports the resulting codebooks and per-edge
indices in a quantized, bit-packed format. Across spline and polynomial KANs,
sampled edge responses exhibit $13$--$35\%$ lower effective rank than their
coefficient representations. Further replicated controls show that
function-space clustering alone is statistically tied with coefficient-space
clustering; the consistent accuracy gain comes from preserving the distinct
sharing structure of the two branches. On a ten-seed MNIST benchmark,
FuncCode compresses spline and GRAM KANs by $31.6\times$ and $17.6\times$
with only $0.31$ and $0.34$\,pp accuracy loss. On a 6.1M-edge convolutional
KAGN, it achieves $19.9\times$ compression while remaining within $0.54$\,pp
of dense accuracy on CIFAR-10 and $1.89$\,pp on CIFAR-100. After compression,
per-edge indices account for up to $99.4\%$ of stored weight bits, making the
representation index-bound. Across nine bit-exact FPGA accelerators,
FuncCode reduces SplineKAN post-route weight memory by $3.87\times$
relative to dense INT4, without increasing cycle count or latency. The FuncCode implementation is available at
\url{https://github.com/OSU-STARLAB/FuncCode}.
\end{abstract}



\section{Introduction}

Kolmogorov--Arnold Networks (KANs) replace each scalar MLP weight with a
learnable univariate function~\cite{liu2024kan}. This gives every connection
its own nonlinear transformation, but it also makes each edge considerably
more expensive to store. Instead of one scalar, an edge contains several
spline, RBF, or polynomial coefficients, often together with a separate base
parameter. For example, a spline-based KAN with a
$784{\to}64{\to}10$ architecture contains $50{,}816$ edges but requires
$457{,}344$ learned parameters, occupying $1{,}786.5$,KiB in FP32. KANs are
therefore compact in edge count but costly in memory, which limits deployment
on memory-constrained hardware.

Existing compression methods address parts of this problem. KAN-specific
quantization reduces the precision of coefficients and activations
~\cite{quantkan,kantize}, while MetaCluster and SHARe-KAN reduce the number of
distinct coefficient vectors through codebook sharing
~\cite{metakan,sharekan}. Hardware-oriented studies further reduce the cost of
evaluating KAN functions or map them to LUT-based accelerators
~\cite{huang2025kanhw,kanele}. These approaches are valuable, but they largely
treat the stored coefficients as the compression target. This overlooks two
properties of a KAN edge. First, different coefficient vectors can represent
similar functions, especially across different basis families. Second, the
basis and base branches capture different components of the edge response and
need not share the same clustering structure.

This observation motivates \funccode{}, a codebook approach that compresses
KANs through their learned edge functions. We evaluate each edge over a fixed
domain to obtain a basis-agnostic function signature and use these signatures
to assign edges to shared codewords. The selected codewords remain in the
original coefficient representation, so the compressed model can be evaluated
without storing sampled functions. We separate the basis and base
branches into independent codebooks, allowing each to retain its own
sharing structure. Controlled experiments show that this branch-aware
representation, rather than the function-space metric alone, provides the most
consistent accuracy benefit.

Compression is useful for deployment only when the resulting representation
reduces the memory stored by the hardware. \funccode{} therefore quantizes and
bit-packs the shared codebooks, scales, and per-edge indices. For a layer with
$E$ edges and a $K$-entry codebook, codebook storage grows with $K$, whereas
index storage grows with $E\lceil\log_2K\rceil$. Because practical KAN layers
typically satisfy $E\gg K$, the compressed model becomes
\emph{index-bound}: most of the remaining weight storage is consumed by narrow
indices rather than repeated coefficient vectors. This predicts that \funccode{} should reduce physical memory without changing the arithmetic schedule, which we test through FPGA synthesis, place-and-route, and RTL co-simulation.

KAN compression should therefore preserve the structure of learned edge
functions and be evaluated in the representation ultimately stored by the
hardware. \funccode{} combines branch-aware function sharing with low-bit
packed deployment to connect software compression directly to hardware memory
savings. Our contributions are threefold:

\begin{itemize}
\item We introduce a basis-agnostic, branch-aware codebook approach for
spline, RBF, and polynomial KANs. Separately coding the basis and base branches
improves accuracy by $4.97$--$28.21$ pp in the controlled ablation. We also
derive the exact storage cost of the quantized codebooks and packed indices.

\item On the ten-seed MNIST benchmark, \funccode{} compresses spline and GRAM
KANs by $31.6\times$ and $17.6\times$ with only $0.31$ and $0.34$ pp accuracy
loss. On 6.1M-edge convolutional KAGNs, it achieves about $20\times$
compression while remaining within $0.54$ pp of dense accuracy on CIFAR-10
and $1.89$ pp on CIFAR-100.

\item Across nine FPGA accelerators, bit-exact verification confirms that \funccode{} reduces SplineKAN post-route weight memory from 147 to 38 BRAM18 blocks ($3.87\times$) relative to dense INT4, without increasing cycle count or latency.
\end{itemize}

\section{Related Work}

\paragraph{KAN architectures.}
The original spline-based KAN~\cite{liu2024kan} has inspired variants,
including RBF-based FastKAN~\cite{li2024fastkan}, polynomial GRAM/KAGN models
~\cite{drokin2024kanconv}, convolutional KANs~\cite{bodner2024convkan}, and Chebyshev and wavelet formulations ~\cite{ss2024chebyshev,bozorgasl2024wavkan}.  
Although these models use different bases, each represents a logical edge with multiple learned coefficients, which is the structural source of the parameter overhead relative to MLPs~\cite{yu2024fairer}. This shared structure motivates compression across KAN families rather than a particular basis. \funccode{} then uses evaluated edge responses as a common interface.

\paragraph{Quantization and model compression.}
Weight sharing, clustering, vector quantization, and low-bit arithmetic are
widely used to reduce neural-network storage
~\cite{chen2015hashednets,han2016deep,gong2015vecquant,esser2020lsq}. Product
quantization further reduces index cost by factorizing a vector into separately
coded subvectors~\cite{jegou2011pq}. Recent KAN-specific methods adapt these
ideas to functional edges. QuantKAN develops branch-aware quantization-aware
training and post-training quantization across several KAN families
~\cite{quantkan}, while KANtize studies low-bit coefficients, spline outputs,
and tabulated spline evaluation~\cite{kantize}. These approaches reduce the
precision of stored parameters but generally retain a distinct coefficient
vector for every edge. In contrast, \funccode{} first reduces the number of
distinct edge representations through sharing and then quantizes the resulting
codebooks.

\paragraph{KAN weight sharing.}
MetaCluster uses a meta-learner to shape coefficient vectors before clustering
them into a shared codebook~\cite{zhao2025metakan, metakan}. Its motivation is that raw $d_B$-dimensional coefficient vectors resist clustering because distances
concentrate in high dimensions~\cite{beyer1999nn,donoho2000curses}; the
meta-learner supplies a low-dimensional manifold by construction. SHARe-KAN applies a
Gain--Shape--Bias decomposition followed by post-training vector quantization
for cache-efficient inference~\cite{sharekan, gersho1992vq}. These methods are the closest
to \funccode{}, as all three replace repeated edge parameters with shared
representatives. \funccode{} differs in three respects: it defines assignments
from sampled edge responses, separates the basis and base branches into
independent codebooks, and exports the resulting representation as quantized,
bit-packed codebooks and indices. Thus, its focus is not only weight sharing,
but also preserving KAN-specific edge structure in the deployed representation.

\paragraph{Structural and training-time KAN compression.}
A separate line reduces KAN cost by changing the model rather than its stored
representation: Shapley-guided pruning~\cite{fan2025shapkan}, Lipschitz and
$L_{1.5}$ regularizers that bound spline complexity~\cite{li2025lipkan}, and
architectural rank reduction~\cite{ta2025prkan}. 
These require retraining or modify the layer's arithmetic.  \funccode{} keeps the trained architecture and datapath fixed and changes only how edge coefficients are fetched, which makes the FPGA comparison in Section~\ref{sec:hardware} controlled.

\paragraph{KAN hardware.}
Inference on memory-constrained accelerators is frequently bandwidth- rather
than compute-bound~\cite{sze2017efficient, jacob2018quantization}, and off-chip DRAM access costs
orders of magnitude more energy per bit than on-chip
SRAM~\cite{horowitz2014energy}. Existing hardware studies primarily reduce the cost of evaluating KAN functions. Huang et al. combine hardware-aware quantization and mapping with analog compute-in-memory~\cite{huang2025kanhw}, while KANEL'E maps learned KAN functions to LUT-centric FPGA accelerators~\cite{kanele}. \funccode{} addresses a complementary bottleneck: the memory required to store edge parameters.
Rather than modifying the arithmetic operator, we hold the datapath and cycle
schedule fixed and replace dense INT4 coefficients with codebook lookups and
per-edge indices. Section~\ref{sec:hardware} evaluates whether this
representation reduces physical memory resources without increasing latency.

\section{FuncCode}
\label{sec:method}

\begin{figure*}[!ht]
\centering
\includegraphics[width=0.95\textwidth]{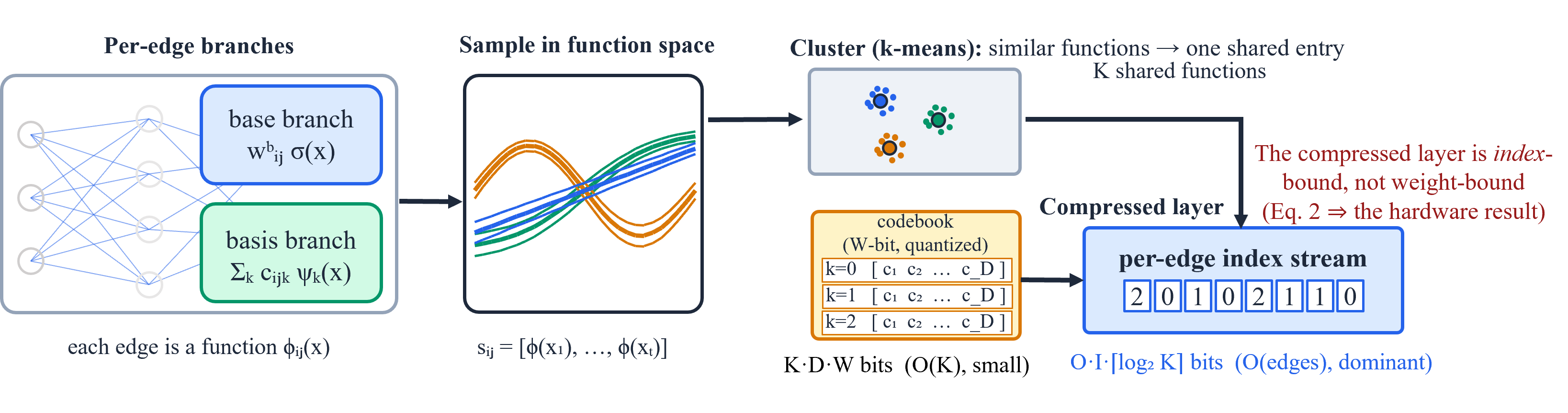}
\caption{\textbf{The \funccode{} pipeline.} Each trained KAN edge splits into
base and basis branches. We sample their responses on a fixed domain to form
function-space signatures $s_{ij}$, cluster similar edge functions with
$k$-means, and store the compressed layer as quantized codebooks plus per-edge
indices. Codebook cost grows with the number of clusters $K$, whereas index
cost grows with the edge count; the layer therefore becomes index-bound, as
Section~\ref{sec:hardware} tests on silicon.}
\label{fig:method}
\end{figure*}

\subsection{KAN layers as edge functions}
\label{sec:kan_edges}

Consider a KAN layer with $I$ input units and $O$ output units; we omit the
layer index for clarity. Whereas an MLP assigns each edge a scalar weight, a
KAN assigns edge $(i,o)$ a learnable univariate function and sums the incoming
responses,
\begin{equation}
u_o
=
\sum_{i=1}^{I} \phi_{oi}(x_i),
\qquad
o=1,\ldots,O,
\label{eq:kan_layer}
\end{equation}
producing $\mathbf{y}=\rho(\mathbf{u})$ with an optional family-specific
transform $\rho$. In SplineKAN and FastKAN, $\rho$ is the identity; in our
GRAM implementation it contains the output normalization and activation
applied after edge accumulation. Each edge function combines a base branch and a basis branch,
\begin{equation}
\phi_{oi}(x_i)
=
w^{\mathrm{base}}_{oi}\,\sigma(x_i)
+
\sum_{k=1}^{d_B}
w^{(k)}_{oi}\,
B_k\!\left(\widetilde{x}_i\right),
\label{eq:edge}
\end{equation}
where $\sigma$ is SiLU, $w^{\mathrm{base}}_{oi}$ is the base weight, and
$\{w^{(k)}_{oi}\}_{k=1}^{d_B}$ are basis coefficients. The transformed input
$\widetilde{x}_i$ captures family-specific preprocessing: it is the identity
for SplineKAN and GRAM, while FastKAN normalizes the input before evaluating
its RBF basis. 

The basis functions $B_k$ differ by family. SplineKAN uses Cox--de~Boor
B-splines with grid size $G$ and degree $p$, giving $d_B=G+p$ coefficients per
edge; FastKAN uses $d_B$ Gaussian radial basis functions; GRAM uses a
polynomial recurrence over a $\tanh$-transformed input, where $d_B$ is one more
than the degree. A layer therefore exposes
$\mathbf{W}^{\mathrm{basis}}\in\mathbb{R}^{O\times I\times d_B}$ and
$\mathbf{W}^{\mathrm{base}}\in\mathbb{R}^{O\times I}$. \funccode{} does not
use the analytical form of $B_k$; it needs only these per-edge parameters and
a routine that evaluates an edge response on a prescribed domain. Signature
construction, clustering, codebook formation, and index storage are therefore
shared across families (Figure~\ref{fig:method}).

\subsection{Function-space codebooks}
\label{sec:funcspace}

We compress each layer independently by replacing its $E=OI$ per-edge
parameter vectors with entries from a shared codebook. The key question is the
space in which assignments are formed. A coefficient-space baseline clusters
$\mathbf{q}_{oi}=[w^{(1)}_{oi},\ldots,w^{(d_B)}_{oi},w^{\mathrm{base}}_{oi}]$.
\funccode{} instead clusters the \emph{functions} those parameters realize:
we evaluate each edge at $T$ points
$\mathcal{Z}=\{z_1,\ldots,z_T\}$ and form
\begin{equation}
\mathbf{s}_{oi}
=
[\phi_{oi}(z_1),\ldots,\phi_{oi}(z_T)]^\top .
\label{eq:function_signature}
\end{equation}
Each signature is standardized to zero mean and unit variance so clustering
emphasizes function shape rather than offset or scale, then $k$-means is
applied in signature space. By default, $T=128$ points span $[-2.5,2.5]$; we
also support domains estimated from layer activations.

The sampled signatures determine assignments only and are never stored at
inference. Each cluster entry is the mean of the original coefficient vectors
assigned to it, so edge $(i,o)$ stores one index $a_{oi}$ and reconstructs as
$\widehat{\mathbf{q}}_{oi}=\mathbf{c}_{a_{oi}}$. We then freeze assignments
and fine-tune the shared entries against the task loss. The coefficient-space
baseline uses the same codebook size, representation, and fine-tuning,
differing only in the space used to form assignments, isolating that choice.

\subsection{Branch-aware codebooks}
\label{sec:branch_aware}

A single codebook gives the whole edge vector one index, forcing the base and
basis branches to vary together. \funccode{} instead compresses them
separately. For the basis branch, we sample
$\sum_{k=1}^{d_B}w^{(k)}_{oi}B_k(x)$ and cluster the resulting signatures in
function space into $K_s$ groups; scalar base weights
$w^{\mathrm{base}}_{oi}$ are clustered into $K_b$ groups. Each edge stores two
indices and reconstructs as
\begin{equation}
\widehat{\mathbf{q}}_{oi}
=
\left[
\mathbf{c}^{s}_{a^{s}_{oi}},
\;
c^{b}_{a^{b}_{oi}}
\right],
\label{eq:branch_reconstruction}
\end{equation}
where $\mathbf{c}^{s}_{j}\in\mathbb{R}^{d_B}$ and
$c^{b}_{j}\in\mathbb{R}$ are basis and base entries. Assignments remain fixed
while both codebooks are fine-tuned. The second index stream is the cost; in
return, each branch keeps its own sharing structure, and
Section~\ref{sec:why} measures why that trade is useful.

\textit{Hardware-aware quantization and export.}
After fixed-assignment fine-tuning, we quantize each codebook vector with
uniform symmetric $b$-bit quantization. For codeword $\mathbf{c}_j$, we use one
per-codeword scale

\begin{equation}
s_j = \|\mathbf{c}_j\|_{\infty}\big/(2^{b-1}\!-\!1), \quad
\mathbf{q}_j = \big[\,\lfloor \mathbf{c}_j/s_j \rceil\,\big]_{\pm(2^{b-1}-1)},
\label{eq:codebook_quantization}
\end{equation}
and reconstruct as $\widehat{\mathbf{c}}_j=s_j\mathbf{q}_j$. Thus each
codeword becomes signed $b$-bit integers plus one scale, contributing the
$32(K_s+K_b)$ term in Equation~\eqref{eq:storage}. We deploy at $b=4$ to match
the W4A4 accelerator datapath. Codebook values and per-edge indices are
bit-packed, with each index using $\lceil\log_2K\rceil$ bits for a $K$-entry
codebook, so reported storage reflects the exported hardware representation
rather than framework tensor types.

\subsection{Exact storage and index dominance}
\label{sec:storage}

With $b$-bit codebooks and one 32-bit scale per entry, a branch-aware layer
with $E=OI$ edges costs
\begin{align}
S_{\mathrm{layer}}
={}&
\underbrace{b(K_s d_B+K_b)}_{\text{codebooks}}
+\underbrace{E\bigl(\lceil\log_2 K_s\rceil+\lceil\log_2 K_b\rceil\bigr)}_{\text{indices}}
\nonumber\\
&+\underbrace{32(K_s+K_b)}_{\text{scales}}\;\text{bits}.
\label{eq:storage}
\end{align}
This is the exact bit count produced by our packed export and used for every
storage result in the paper. Codebook and scale terms grow with cluster count,
whereas the index term grows with the number of edges and shrinks only
logarithmically in $K$. Whenever $E\gg K_s,K_b$, as in every layer worth
compressing, indices dominate. For the spline model at
$(K_s,K_b)=(32,16)$, the two indices cost nine bits per edge and account for
$98.9\%$ of deployed weight storage. \funccode{} therefore turns a
coefficient-heavy layer into an \emph{index-bound} one, a prediction tested
directly in Section~\ref{sec:hardware}.

\section{Why Function Space?}
\label{sec:why}

\textit{Effective rank.} We first ask whether trained edge functions are more redundant in function space than their parameters suggest. For each layer, we form a signature matrix $\mathbf{S}\in\mathbb{R}^{E\times T}$ whose rows are normalized edge
signatures, and a coefficient matrix
$\mathbf{Q}\in\mathbb{R}^{E\times(d_B+1)}$ whose rows are the corresponding
edge parameters. After column-centering, we measure effective rank from the
entropy of normalized singular values,


{\small
\begin{equation}
\operatorname{erank}(\mathbf{X}) = \exp\!\left(-\sum_j p_j\log p_j\right),
\qquad p_j=\frac{\sigma_j}{\sum_\ell \sigma_\ell},
\label{eq:effective_rank}
\end{equation}
}

where $\{\sigma_j\}$ are the singular values of $\mathbf{X}$, and report
$\operatorname{erank}(\mathbf{S})/\operatorname{erank}(\mathbf{Q})$ averaged
over layers.

\begin{table}[!ht]
\centering
\caption{Effective-rank ratio between function and coefficient spaces.
A value below one means the functions occupy fewer effective
directions than their coefficient vectors.}
\label{tab:rank}
\small
\begin{tabular}{lrr}
\toprule
Dataset & SplineKAN & GRAM \\
\midrule
MNIST         & 0.822 & \textbf{0.666} \\
Fashion-MNIST & 0.831 & \textbf{0.658} \\
CIFAR-10      & 0.863 & \textbf{0.655} \\
Wine          & 0.857 & \textbf{0.653} \\
\bottomrule
\end{tabular}
\end{table}

Every ratio in Table~\ref{tab:rank} lies below one, with stable ordering across
four datasets of different difficulty, suggesting that the redundancy is
basis-dependent rather than task-specific. GRAM concentrates more strongly
than SplineKAN. The gap is meaningful but not a collapse: a ratio of $0.82$
is an $18\%$ reduction, not evidence of a low-dimensional manifold. We omit FastKAN, whose ratio depends on whether its input normalization is bypassed before sampling ($0.118$ against $0.777$); the appendix reports both.

\textit{Clustering confirms the rank advantage.}
We next hold codebook size, storage, and fine-tuning fixed and vary only the
clustering space. On SplineKAN with MNIST and W4 codebooks, function-space
clustering improves accuracy by $0.76$, $0.63$, $0.25$, and $0.31$ percentage
points at $K\in\{16,32,64,128\}$. The gain appears at every size and is largest
when the codebook is tightest, as the rank analysis predicts: with few
representatives, grouping edges by the functions they realize preserves more
useful structure than grouping them by coefficient proximity.

\textit{The branches carry complementary information.}
Should the base branch enter a shared-edge signature at all? We compare two
otherwise identical SplineKAN runs, one sampling the complete edge function
and one sampling only the basis branch, under W4 quantization.

\begin{table}[!ht]
\centering
\caption{Effect of including the base branch in the signature.
SplineKAN on MNIST; all settings identical except signature construction.}
\label{tab:sig}
\small
\begin{tabular}{lrrr}
\toprule
$K$ & Base included & Base excluded & $\Delta$ (pp) \\
\midrule
16  & \textbf{92.63} & 64.42 & \textbf{+28.21} \\
32  & \textbf{93.68} & 83.39 & +10.29 \\
64  & \textbf{94.37} & 87.73 & +6.64 \\
128 & \textbf{95.16} & 90.19 & +4.97 \\
\bottomrule
\end{tabular}
\end{table}

Including the base response helps at every codebook size, with the gap widening
from $4.97$ to $28.21$ percentage points as $K$ falls. When the codebook is
small, assigning a full edge from a signature that omits one branch loses
important information. This motivates separating the branches: rather than
forcing one shared assignment to capture both, branch-aware coding lets each
retain its own sharing structure.




\textit{The gap is primarily representational, not an optimization artifact.}
A natural objection is that the loss comes from greedy hard $k$-means rather than
the representation. We therefore replace fixed assignments with a differentiable
soft-to-hard relaxation, trained jointly and hardened at export, at identical
storage.

\begin{table}[!ht]
\centering
\caption{Hard versus learned assignments for the index-efficient variant, which
keys both branches from one index. SplineKAN, MNIST, W4, seed~42.}
\label{tab:soft}
\small
\begin{tabular}{lrrr}
\toprule
$K$ & Hard $k$-means & Soft-to-hard & $\Delta$ (pp) \\
\midrule
16 & 74.76 & \textbf{78.13} & +3.37 \\
32 & 84.93 & 84.93 & 0.00 \\
\bottomrule
\end{tabular}
\end{table}

Learning the assignment gains $3.37$\,pp at $K{=}16$ and nothing at $K{=}32$,
yet stays $17.19$ and $10.39$\,pp below branch-aware $(32,16)$ at $95.32\%$.
At $K{=}32$, the branch-aware and index-efficient variants provide
$31.6\times$ and $56.2\times$ compression, respectively. Replication does not
preserve the sign: across four three-seed configurations, the mean learned-
assignment result is below hard $k$-means at W4 in every case, by
$0.39$--$1.22$\,pp. The dominant limitation is therefore representational:
one index retains only $K$ coupled branch combinations, rather than allowing
the two branches to select their representatives independently.

\section{Results and Analysis}
\label{sec:results}

The controlled benchmark trains spline, FastKAN, GRAM, and an MLP reference on
MNIST, CIFAR-10, and CIFAR-100 over ten seeds, with W4 codebooks and bit-exact
storage. Controls use three or five seeds as noted. We additionally evaluate a
30.6M-parameter, 6.1M-edge convolutional KAGN on CIFAR-10/100 over three seeds.

\begin{table}[!ht]
\centering
\caption{\textbf{MNIST controlled benchmark, ten seeds, W4 codebooks}
(mean\,$\pm$\,std). All four families share one protocol, so the comparison
across bases and codebook arms is controlled. Compression is against each
family's own dense FP32 model. LSQ-MLP is a learned-step-size QAT baseline with
a larger (30-epoch) budget, reported with that caveat. The flattened CIFAR
stress tables are in the Appendix; Section~\ref{sec:conv} reports CIFAR on
convolutional backbones.}
\label{tab:main}
\small
\setlength{\tabcolsep}{4pt}
\begin{tabular}{llcc}
\toprule
Family & Method & Acc.\ (\%) & Comp. \\
\midrule
spline & dense FP32           & 95.86 $\pm$ 0.20 & 1.0$\times$ \\
spline & coefficient $K{=}32$ & 93.22 $\pm$ 0.53 & 56.6$\times$ \\
spline & function $K{=}32$    & 92.96 $\pm$ 1.47 & 56.6$\times$ \\
spline & \textbf{branch $(32,16)$} & \textbf{95.55 $\pm$ 0.31} & 31.6$\times$ \\
\midrule
FastKAN & dense FP32           & 96.83 $\pm$ 0.22 & 1.0$\times$ \\
FastKAN & coefficient $K{=}32$ & \textbf{95.42 $\pm$ 0.38} & 56.6$\times$ \\
FastKAN & function $K{=}32$    & 86.83 $\pm$ 11.70 & 56.6$\times$ \\
FastKAN & branch $(32,16)$     & 93.85 $\pm$ 1.26 & 31.6$\times$ \\
\midrule
GRAM & dense FP32           & 96.74 $\pm$ 0.08 & 1.0$\times$ \\
GRAM & coefficient $K{=}32$ & 95.60 $\pm$ 0.27 & 31.6$\times$ \\
GRAM & function $K{=}32$    & 95.46 $\pm$ 0.27 & 31.6$\times$ \\
GRAM & \textbf{branch $(32,16)$} & \textbf{96.40 $\pm$ 0.21} & 17.6$\times$ \\
\midrule
MLP & dense FP32      & 96.06 $\pm$ 0.19 & 1.0$\times$ \\
MLP & uniform W4 PTQ  & 95.66 $\pm$ 0.24 & 8.0$\times$ \\
MLP & LSQ QAT W4      & 97.36 $\pm$ 0.09 & 8.0$\times$ \\
\bottomrule
\end{tabular}
\end{table}

\subsection{Accuracy--storage trade-off}
On MNIST, branch-aware \funccode{} retains spline within $0.31$\,pp of dense at
$31.6\times$ compression and GRAM within $0.34$\,pp at $17.6\times$, with
small seed spreads. The sharper test is matched storage: the spline model uses
roughly the same $56$\,KiB as the best small dense KAN we could train, yet
reaches $95.55\%$ rather than $92.25\%$ ($+3.3$\,pp). Thus the gain is not
simply fewer stored parameters. \funccode{} retains the width and edge-function
capacity of the larger network while amortizing those functions through a
shared dictionary. The MLP comparison makes a different point: W4 PTQ gives
$8\times$ compression with a $0.40$\,pp loss, whereas spline \funccode{}
reaches $31.6\times$ with a $0.31$\,pp loss. We therefore claim greater
\emph{compressibility} of the KAN representation, not a smaller absolute model.
The flattened CIFAR runs in the Appendix are a stress test rather than a vision
benchmark; absolute accuracy there is limited by the lack of convolutions.

\begin{figure}[!ht]
\centering
\includegraphics[width=\linewidth]{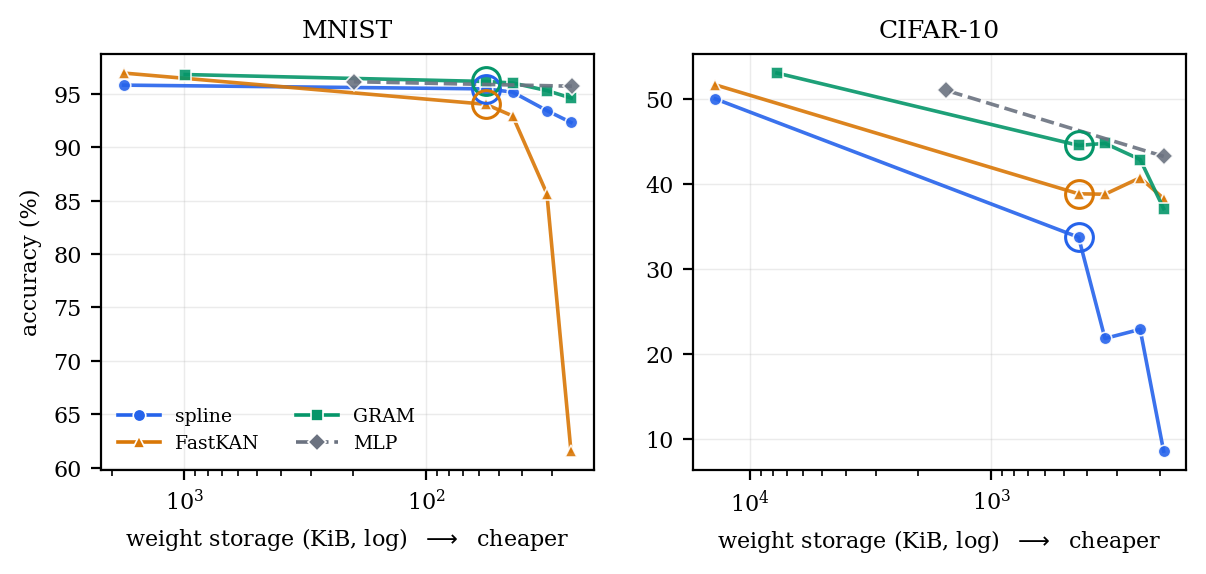}
\caption{\textbf{Accuracy versus storage, ten seeds.} Each KAN curve runs from
its dense model through branch-aware $(16,8)$ and $(32,16)$; the hollow ring
marks the deployed $(32,16)$ point. The MLP (grey, dashed) shows dense and
uniform W4 PTQ; the star is the LSQ-QAT MLP, the strongest compressed model on
CIFAR-10 (with a larger training budget; see the baselines discussion).
Storage is bit-exact and the axis is reversed, so cheaper is to the right.}
\label{fig:pareto}
\end{figure}

\subsection{Scaling to convolutional KANs}
\label{sec:conv}

We next remove that architectural confound with an 8-layer convolutional KAGN
(30.6M parameters, 6.1M edges, Gram degree 3), which reaches
$91.43\pm0.21\%$ on CIFAR-10 and $62.83\pm0.96\%$ on CIFAR-100. This is also
a scale test for the method: one late layer contains 1.77M edges, so an
explicit $[E,T]$ signature matrix would require roughly $900$\,MB. We instead
use the exact identity $G=B^\top B=LL^\top$: Euclidean $k$-means on
$L^\top\mathbf{w}$ is function-space $k$-means under the basis-induced
metric. The resulting five-dimensional clustering path uses $26\times$ less
memory and matches the explicit metric to $3.4\times10^{-7}$. At 6.1M edges,
index bits account for $99.4\%$ of compressed storage, extending the
index-bound argument beyond the small fully connected models. Codebook storage
is therefore negligible; all convolutional results use W8 because W4 is
empirically unstable on this track.

\begin{table}[!ht]
\centering
\caption{\textbf{Convolutional KAGN, three seeds, w8 codebooks.} Bits per edge
(b/e) include codebooks, scales, masks, and all normalization parameters;
compression is against the FP32 dense model ($160.0$\,b/e). Iso-dense is a
width-narrowed dense net at matched storage (one seed).}
\label{tab:conv}
\footnotesize
\setlength{\tabcolsep}{2.2pt}
\begin{tabular}{lcccc}
\toprule
Method & b/e & Comp. & CIFAR-10 & CIFAR-100 \\
\midrule
dense FP32          & 160.0 & 1.0$\times$  & 91.43 $\pm$ 0.21 & 62.83 $\pm$ 0.96 \\
iso-dense           & 5.03  & 31.8$\times$ & 77.29 & 32.65 \\
\funccode{} $K{=}32$  & 5.03  & 31.8$\times$ & 90.25 $\pm$ 0.18 & 57.65 $\pm$ 0.64 \\
\funccode{} $K{=}64$  & 6.03  & 26.5$\times$ & 90.50 $\pm$ 0.41 & 59.48 $\pm$ 1.02 \\
\funccode{} $K{=}256$ & 8.05  & 19.9$\times$ & 90.89 $\pm$ 0.32 & 60.94 $\pm$ 0.79 \\
prune 90\% $+$ W4   & 7.04  & 22.7$\times$ & 90.82 $\pm$ 0.19 & 60.89 $\pm$ 1.11 \\
LSQ QAT W2          & 10.04 & 15.9$\times$ & 90.98 $\pm$ 0.17 & 61.34 $\pm$ 0.76 \\
\bottomrule
\end{tabular}
\end{table}

At $K{=}256$, \funccode{} is within $0.54\pm0.39$\,pp of dense on CIFAR-10 and
$1.89\pm1.24$\,pp on CIFAR-100 at about $20\times$ compression. The
matched-storage comparison remains favorable at this scale: at five bits per
edge, compressing the full model beats a width-reduced dense network by
$12.96$ and about $25$\,pp. This reproduces the MNIST conclusion in a model
with two orders of magnitude more edges---the useful capacity resides in the
large network, and sharing is a better use of the bit budget than shrinking
it. A sensitivity ablation points to a practical refinement: leaving only the
first convolution and classifier head uncompressed improves CIFAR-100 by
$2.59$\,pp for $6.6\%$ more storage (one seed). Those layers are only $0.4\%$
of the edges but appear disproportionately fragile, so we treat selective
exemption as a deployment option rather than a change to the core method.

Three caveats are material. The whitening metric reduces function-space error
but does not improve accuracy over coefficient clustering in the single-seed
CIFAR-100 control. The planned 4-layer CIFAR-10 backbone reached only
$64.26\%$ and was replaced by the prespecified 8-layer variant. Finally, the
CIFAR-100 dense model is $3.17$\,pp below its $66\%$ target; relative
comparisons remain controlled because all compression arms share the same
checkpoint and fine-tuning budget.

\subsection{What drives the gain?}

\textit{Metric versus representation.} At matched $K$ and storage, five-seed controls do \emph{not} show a reliable single-codebook advantage for function-space over coefficient-space clustering: all six paired confidence intervals include zero, and removing signature normalization does not materially change the result. This corrects the weaker single-seed trend seen in the exploratory sweep. The robust gain is instead the \emph{representation}: branch-aware sharing. On MNIST, spline reaches $95.55\%$ with two branch codebooks versus $93.22\%$ for the best single-codebook arm; the corresponding flattened CIFAR-10 gap is $5.23$\ pp. GRAM shows the same qualitative advantage on MNIST and CIFAR-100. This is consistent with Tables~\ref{tab:sig} and~\ref{tab:ablate}: the base and basis branches carry complementary information, so a single assignment must explain two different sources of variation. Function-space signatures remain a natural, family-agnostic way to define similarity; the data say that branch separation, not the metric alone, is what consistently preserves accuracy.

\textit{Indices and assignments.} Huffman coding the learned index streams saves only $3.1$--$5.6\%$ of total storage ($\eta\approx0.9$--$0.99$), so fixed-width indices are already close to the entropy bound. Learned soft-to-hard assignments also fail to beat hard $k$-means across the replicated configurations. The remaining loss is therefore primarily representational, a matter of how many shared functions are available, rather than an artifact of index coding or discrete optimization.

\subsection{Competing sharing and compression baselines}

Under FP32 fine-tuning, MetaCluster and GSB are competitive, but at 4-bit
deployment their GRAM accuracies fall by $17.94$ and $15.74$\,pp, respectively,
versus $0.20$\,pp for branch-aware \funccode{}. The valid FastKAN MetaCluster
arm is more robust, so this is a family-dependent low-bit effect rather than a
universal failure of prior sharing methods.

\begin{table}[!ht]
\centering
\caption{Sharing baselines under 4-bit deployment quantization
(GRAM, flattened CIFAR-10, width 64, five seeds, QAT numbers; PTQ is worse
for every baseline). GSB is evaluated at storage matched to branch-aware
\funccode{} ($\approx 217$\,KiB, nine bits per edge); its FP32 model is only
$2.3\times$ compressed.}
\label{tab:shared}
\footnotesize
\setlength{\tabcolsep}{2.4pt}
\begin{tabular}{lccc}
\toprule
Method & FP32-FT & QAT 4-bit & Comp. \\
\midrule
\funccode{} branch $(32,16)$ & 47.06 $\pm$ 0.70 & \textbf{46.86 $\pm$ 0.85} & 17.7$\times$ \\
\funccode{} coeff.\ $K{=}32$ & 47.00 $\pm$ 0.61 & \textbf{46.93 $\pm$ 0.73} & 31.9$\times$ \\
MetaCluster $K{=}32$         & \textbf{48.70 $\pm$ 0.85} & 30.76 $\pm$ 6.38 & 31.9$\times$ \\
MetaCluster $K{=}16$         & 48.38 $\pm$ 0.71 & 28.96 $\pm$ 10.46 & 39.9$\times$ \\
GSB $K{=}32$ (matched)       & \textbf{49.44 $\pm$ 0.60} & 33.70 $\pm$ 3.16 & 17.7$\times$ \\
\bottomrule
\end{tabular}
\end{table}
Two stronger controls define the boundary more clearly. Rank-2 low-rank
factorization essentially matches \funccode{} on MNIST
($95.60\pm0.69\%$ versus $95.55\pm0.31\%$) and is higher on flattened
CIFAR-10, but its W4 frontier is unstable and non-monotone as rank grows. More
fundamentally, low rank changes the layer into two dense products, whereas
\funccode{} preserves the original arithmetic and changes only how edge
coefficients are fetched. LSQ-QAT likewise makes the MLP the strongest
compressed model on flattened CIFAR, but uses a larger 30-epoch QAT budget and
addresses a different model family. These controls rule out an absolute
accuracy-per-bit claim. They also sharpen what the FPGA experiment must show:
if the value of \funccode{} is its memory representation, the measured gain
should appear as lower BRAM at the same arithmetic schedule and latency. Under Gaussian noise (severity $0.4$), branch-aware spline stays within
$1.5$\,pp of its dense parent ($86.41$ versus $87.92$), whereas the two
single-codebook arms fall to $78.23$ and $67.79$; dropout corruption shows the
same ordering. 

\subsection{Deployment ablations}
W4 is effectively free for the fully connected branch-aware models: no row
loses more than $0.30$\,pp from W8. W2 is qualitatively different, costing
$1.2$--$7.3$\,pp for branch-aware models and collapsing the single-codebook
FastKAN arm to $14.67\%$. The asymmetry is informative: separate dictionaries
can absorb quantization error independently, while a single codebook must
represent the entire edge with one quantized prototype. We therefore use W4
for the FPGA path, where it also matches the activation width. The large
convolutional study uses W8 because its codebook bits are almost free in
storage and W4 exhibits a $9.99$\,pp run-to-run spread.
\begin{table}[!ht]
\centering
\caption{\textbf{Codebook precision sweep} (MNIST, 3 seeds, branch-aware
$(32,16)$ and function-only $K{=}32$). W4 is nearly free; W2 is not. The
collapse arrives earlier for function-only codebooks, which have fewer entries
absorbing the same quantization error.}
\label{tab:bits}
\small
\setlength{\tabcolsep}{4.5pt}
\begin{tabular}{llrrrr}
\toprule
Family & Method & W8 & W6 & W4 & W2 \\
\midrule
spline  & branch $(32,16)$ & 95.57 & 95.57 & \textbf{95.44} & 94.25 \\
spline  & function $(32)$  & 93.85 & 93.88 & 93.41 & 73.42 \\
GRAM    & branch $(32,16)$ & 96.43 & 96.40 & \textbf{96.13} & 90.29 \\
GRAM    & function $(32)$  & 95.50 & 95.48 & 95.27 & 74.41 \\
FastKAN & branch $(32,16)$ & 94.28 & 94.30 & \textbf{94.00} & 86.70 \\
FastKAN & function $(32)$  & 93.66 & 93.61 & 85.64 & 14.67 \\
\bottomrule
\end{tabular}
\end{table}

The codebook sweep places the knee at $(K_s,K_b)=(32,16)$: reducing the base
codebook to four entries collapses accuracy, while increasing to $(64,16)$
adds only $0.10$\,pp. Forcing both branches through one index is worse still
($84.93\%$ versus $95.32\%$), confirming that the second narrow index stream
is a necessary cost. We therefore synthesize branch-aware $(32,16)$ at W4.

\begin{table}[!ht]
\centering
\caption{Ablations (SplineKAN, MNIST, W4, seed 42). The
$(K_s,K_b)$ grid shows a knee at $(32,16)$; the sparse residual branch (SRB)
trades storage back for accuracy; index-efficient sharing, which forces one
index to key \emph{both} branches, is a controlled failure.}
\label{tab:ablate}
\small
\begin{tabular}{llrr}
\toprule
Variant & Setting & Acc.\ (\%) & Comp. \\
\midrule
branch    & $(16,4)$  & 88.61 & 47.64$\times$ \\
branch    & $(16,8)$  & 94.72 & 40.84$\times$ \\
branch    & $(32,16)$ & \textbf{95.32} & 31.64$\times$ \\
branch    & $(64,16)$ & 95.42 & 28.28$\times$ \\
\midrule
SRB       & $K{=}32,\rho{=}0.10$ & 91.77 & 40.97$\times$ \\
SRB       & $K{=}32,\rho{=}0.50$ & 94.82 & 19.73$\times$ \\
\midrule
index-eff & $K{=}32$  & 84.93 & 56.18$\times$ \\
\bottomrule
\end{tabular}
\end{table}

\section{Hardware}
\label{sec:hardware}

The storage model predicts an index-bound accelerator: replacing dense INT4
weights by indices plus codebooks should reduce memory without changing the
arithmetic schedule. We test nine accelerators---spline, GRAM, and convolutional
KAGN, each as FP32, INT4 LSQ, and \funccode{} W4A4. Within each family the two
quantized designs share the same loop nest, II=1 pipeline, unrolling, target
clock, and testbench; only the weight fetch changes from a dense ROM read to
index plus codebook lookup. This controls the comparison more tightly than a
cross-design throughput benchmark.

The arithmetic contract is frozen across the QAT graph, NumPy fixed-point
emulator, and HLS C++. All six quantized accelerators match emulator top-1 on
the full 10k MNIST test set, match INT32 logits bit-exactly on 1,000 C-simulation
vectors, and reproduce those outputs bit-exactly in RTL co-simulation. FP32
variants agree within floating-point tolerance with exact argmax, and every
routable design meets timing. Thus the resource numbers below describe the
same deployed arithmetic whose accuracy is reported, rather than an analytical
proxy.

\begin{table}[!ht]
\centering
\caption{Nine accelerators on MNIST. Accuracy is on the full 10k test
set under \emph{deployed} arithmetic (L1-verified), so it is not directly
comparable to Table~\ref{tab:main}'s training-time numbers. BRAM18 and latency
are post-synthesis (\texttt{xczu9eg}); post-route BRAM18 on the
\texttt{xczu7ev} is in parentheses. Within each family the two quantized
designs have \emph{identical} cycle counts.}
\label{tab:hw}
\small
\setlength{\tabcolsep}{3.5pt}
\begin{tabular}{llrrrr}
\toprule
Family & Design & Acc.\% & KiB & BRAM18 & Lat.\,$\mu$s \\
\midrule
Spline & FP32 dense       & 96.28 & 1786.5 & 842 ($-$)        & 7086.6 \\
Spline & LSQ W4A4         & 96.22 & 223.3  & 108 (147)        & 340.1 \\
Spline & \funccode{}      & 95.22 & 56.5   & \textbf{32} (38) & 340.1 \\
\midrule
GRAM   & FP32 dense       & 96.68 & 992.5  & 469 (608)        & 1933.9 \\
GRAM   & LSQ W4A4         & \textbf{97.23} & 124.1 & 76 (90)  & 343.6 \\
GRAM   & \funccode{}      & 96.09 & 56.3   & \textbf{43} (46) & 343.6 \\
\midrule
Conv   & FP32 dense       & 95.43 & 99.1   & 96 (118)         & 12525.3 \\
Conv   & LSQ W4A4         & 92.74 & 12.4   & 41 (38)          & 1963.1 \\
Conv   & \funccode{}      & 89.78 & 7.5    & \textbf{36} (32) & 1963.1 \\
\bottomrule
\end{tabular}
\end{table}

For spline, INT4 reduces post-synthesis BRAM18 from 842 to 108; changing only
the weight format to \funccode{} lowers it to 32, or $3.4\times$ beyond dense
INT4 ($3.87\times$ post-route), while preserving the same 50,983 cycles and
$340.1\,\mu$s latency. GRAM shows the same signature ($1.96\times$ post-route
BRAM reduction at unchanged latency). The convolutional accelerator gains less
($1.19\times$ post-route), consistent with its smaller coefficient dimension
and existing spatial sharing.

\begin{table}[!ht]
\centering
\caption{The win decomposes into two independent causes. The first
stage is a precision change; the second is a weight-memory format change and
nothing else. The signature of the second stage is that storage and BRAM fall
while latency is \emph{exactly} unchanged.}
\label{tab:stages}
\small
\setlength{\tabcolsep}{4pt}
\begin{tabular}{lrrr}
\toprule
& Spline & GRAM & Conv \\
\midrule
\multicolumn{4}{l}{\emph{FP32 $\rightarrow$ INT4 (precision)}} \\
\quad weight storage      & 8.0$\times$ & 8.0$\times$ & 8.0$\times$ \\
\quad BRAM18 (syn)        & 7.8$\times$ & 6.2$\times$ & 2.3$\times$ \\
\quad latency             & 20.8$\times$ & 5.6$\times$ & 6.4$\times$ \\
\quad accuracy cost (pp)  & 0.06 & $-0.55$ & 2.69 \\
\midrule
\multicolumn{4}{l}{\emph{INT4 $\rightarrow$ \funccode{} (memory format)}} \\
\quad weight storage      & \textbf{3.95}$\times$ & 2.20$\times$ & 1.65$\times$ \\
\quad BRAM18 (syn)        & 3.4$\times$ & 1.77$\times$ & 1.14$\times$ \\
\quad BRAM18 (route)      & \textbf{3.87}$\times$ & 1.96$\times$ & 1.19$\times$ \\
\quad latency             & \textbf{1.00}$\times$ & \textbf{1.00}$\times$ & \textbf{1.00}$\times$ \\
\quad accuracy cost (pp)  & 1.00 & 1.14 & 2.96 \\
\midrule
composite BRAM vs FP32    & \textbf{26.3}$\times$ & 10.9$\times$ & 2.7$\times$ \\
index share of storage    & 98.9\% & 99.1\% & 82.3\% \\
\bottomrule
\end{tabular}
\end{table}

Table~\ref{tab:stages} isolates the contribution. FP32$\rightarrow$INT4 is a
precision change and produces the expected arithmetic speedup; INT4$\rightarrow$
\funccode{} is a memory-format change and leaves the compute schedule untouched.
Its diagnostic signature is therefore lower storage and BRAM with exactly
$1.00\times$ latency, which appears in all three families. For spline, the
analytical $3.95\times$ storage reduction becomes $3.4$--$3.87\times$ in BRAM,
with the residual explained by discrete BRAM tiling. The smaller convolutional
gain is likewise consistent with its lower coefficient dimension and existing
spatial sharing. 



\section{Limitations}

This study focuses on representation-level compression and deployed memory.
Our results show that branch-aware sharing, rather than the clustering metric
alone, provides the most consistent accuracy benefit. Sensitivity varies across
KAN families and under aggressive compression, suggesting that adaptive
codebook allocation may provide further gains. The FPGA evaluation isolates
the memory-format effect under a fixed arithmetic schedule; broader validation
on larger workloads, batched inference, and measured energy remains future
work.



\section{Conclusion}

\funccode{} compresses KANs by sharing learned edge functions through
branch-aware codebooks and exporting them as low-bit codewords and narrow
indices. On ten-seed MNIST, it compresses spline and GRAM KANs by
$31.6\times$ and $17.6\times$ with only $0.31$ and $0.34$,pp accuracy loss.
On a 6.1M-edge convolutional KAGN, it achieves about $20\times$ compression
within $0.54$,pp of dense accuracy on CIFAR-10 and $1.89$,pp on CIFAR-100.
On FPGA, the same representation reduces SplineKAN post-route weight memory by
$3.87\times$ relative to dense INT4 without increasing latency. These results
show that structured function sharing can translate directly into practical
hardware memory savings.


\bibliography{aaai2027}

@misc{liu2024kan,
      title={KAN: Kolmogorov-Arnold Networks}, 
      author={Ziming Liu and Yixuan Wang and Sachin Vaidya and Fabian Ruehle and James Halverson and Marin Soljačić and Thomas Y. Hou and Max Tegmark},
      year={2025},
      eprint={2404.19756},
      archivePrefix={arXiv},
      primaryClass={cs.LG},
      url={https://arxiv.org/abs/2404.19756}, 
}

@misc{li2024fastkan,
      title={Kolmogorov-Arnold Networks are Radial Basis Function Networks}, 
      author={Ziyao Li},
      year={2024},
      eprint={2405.06721},
      archivePrefix={arXiv},
      primaryClass={cs.LG},
      url={https://arxiv.org/abs/2405.06721}, 
}

@misc{ss2024chebyshev,
      title={Chebyshev Polynomial-Based Kolmogorov-Arnold Networks: An Efficient Architecture for Nonlinear Function Approximation}, 
      author={Sidharth SS and Keerthana AR and Gokul R and Anas KP},
      year={2024},
      eprint={2405.07200},
      archivePrefix={arXiv},
      primaryClass={cs.LG},
      url={https://arxiv.org/abs/2405.07200}, 
}

@misc{bozorgasl2024wavkan,
      title={Wav-KAN: Wavelet Kolmogorov-Arnold Networks}, 
      author={Zavareh Bozorgasl and Hao Chen},
      year={2024},
      eprint={2405.12832},
      archivePrefix={arXiv},
      primaryClass={cs.LG},
      url={https://arxiv.org/abs/2405.12832}, 
}

@misc{drokin2024kanconv,
      title={Kolmogorov-Arnold Convolutions: Design Principles and Empirical Studies}, 
      author={Ivan Drokin},
      year={2024},
      eprint={2407.01092},
      archivePrefix={arXiv},
      primaryClass={cs.CV},
      url={https://arxiv.org/abs/2407.01092}, 
}

@misc{han2016deep,
      title={Deep Compression: Compressing Deep Neural Networks with Pruning, Trained Quantization and Huffman Coding}, 
      author={Song Han and Huizi Mao and William J. Dally},
      year={2016},
      eprint={1510.00149},
      archivePrefix={arXiv},
      primaryClass={cs.CV},
      url={https://arxiv.org/abs/1510.00149}, 
}

@misc{chen2015hashednets,
      title={Compressing Neural Networks with the Hashing Trick}, 
      author={Wenlin Chen and James T. Wilson and Stephen Tyree and Kilian Q. Weinberger and Yixin Chen},
      year={2015},
      eprint={1504.04788},
      archivePrefix={arXiv},
      primaryClass={cs.LG},
      url={https://arxiv.org/abs/1504.04788}, 
}

@misc{esser2020lsq,
      title={Learned Step Size Quantization}, 
      author={Steven K. Esser and Jeffrey L. McKinstry and Deepika Bablani and Rathinakumar Appuswamy and Dharmendra S. Modha},
      year={2020},
      eprint={1902.08153},
      archivePrefix={arXiv},
      primaryClass={cs.LG},
      url={https://arxiv.org/abs/1902.08153}, 
}

@misc{gong2015vecquant,
      title={Compressing Deep Convolutional Networks using Vector Quantization}, 
      author={Yunchao Gong and Liu Liu and Ming Yang and Lubomir Bourdev},
      year={2014},
      eprint={1412.6115},
      archivePrefix={arXiv},
      primaryClass={cs.CV},
      url={https://arxiv.org/abs/1412.6115}, 
}

@misc{kantize,
      title={KANtize: Exploring Low-bit Quantization of Kolmogorov-Arnold Networks for Efficient Inference}, 
      author={Sohaib Errabii and Olivier Sentieys and Marcello Traiola},
      year={2026},
      eprint={2603.17230},
      archivePrefix={arXiv},
      primaryClass={cs.AR},
      url={https://arxiv.org/abs/2603.17230}, 
}

@misc{sharekan,
      title={SHARe-KAN: Post-Training Vector Quantization for Cache-Resident KAN Inference}, 
      author={Jeff Smith},
      year={2026},
      eprint={2512.15742},
      archivePrefix={arXiv},
      primaryClass={cs.LG},
      url={https://arxiv.org/abs/2512.15742}, 
}

@inproceedings{kanele,
   title={KANELÉ: Kolmogorov–Arnold Networks for Efficient LUT-based Evaluation},
   url={http://dx.doi.org/10.1145/3748173.3779202},
   DOI={10.1145/3748173.3779202},
   booktitle={Proceedings of the 2026 ACM/SIGDA International Symposium on Field Programmable Gate Arrays},
   publisher={ACM},
   author={Hoang, Duc and Gupta, Aarush and Harris, Philip C},
   year={2026},
   month=Feb, pages={44–55} }

@misc{huang2025kanhw,
      title={Hardware Acceleration of Kolmogorov-Arnold Network (KAN) in Large-Scale Systems}, 
      author={Wei-Hsing Huang and Jianwei Jia and Yuyao Kong and Faaiq Waqar and Tai-Hao Wen and Meng-Fan Chang and Shimeng Yu},
      year={2025},
      eprint={2509.05937},
      archivePrefix={arXiv},
      primaryClass={cs.AR},
      url={https://arxiv.org/abs/2509.05937}, 
}

@misc{metakan,
      title={MetaCluster: Enabling Deep Compression of Kolmogorov-Arnold Network}, 
      author={Matthew Raffel and Adwaith Renjith and Lizhong Chen},
      year={2026},
      eprint={2510.19105},
      archivePrefix={arXiv},
      primaryClass={cs.LG},
      url={https://arxiv.org/abs/2510.19105}, 
}

@misc{quantkan,
      title={QuantKAN: A Unified Quantization Framework for Kolmogorov Arnold Networks}, 
      author={Kazi Ahmed Asif Fuad and Lizhong Chen},
      year={2026},
      eprint={2511.18689},
      archivePrefix={arXiv},
      primaryClass={cs.LG},
      url={https://arxiv.org/abs/2511.18689}, 
}

@ARTICLE{jegou2011pq,
  author={Jégou, Herve and Douze, Matthijs and Schmid, Cordelia},
  journal={IEEE Transactions on Pattern Analysis and Machine Intelligence}, 
  title={Product Quantization for Nearest Neighbor Search}, 
  year={2011},
  volume={33},
  number={1},
  pages={117-128},
  doi={10.1109/TPAMI.2010.57}}

@misc{bodner2024convkan,
      title={Convolutional Kolmogorov-Arnold Networks}, 
      author={Alexander Dylan Bodner and Antonio Santiago Tepsich and Jack Natan Spolski and Santiago Pourteau},
      year={2025},
      eprint={2406.13155},
      archivePrefix={arXiv},
      primaryClass={cs.CV},
      url={https://arxiv.org/abs/2406.13155}, 
}

@misc{yu2024fairer,
      title={KAN or MLP: A Fairer Comparison}, 
      author={Runpeng Yu and Weihao Yu and Xinchao Wang},
      year={2024},
      eprint={2407.16674},
      archivePrefix={arXiv},
      primaryClass={cs.LG},
      url={https://arxiv.org/abs/2407.16674}, 
}

@misc{zhao2025metakan,
      title={Improving Memory Efficiency for Training KANs via Meta Learning}, 
      author={Zhangchi Zhao and Jun Shu and Deyu Meng and Zongben Xu},
      year={2025},
      eprint={2506.07549},
      archivePrefix={arXiv},
      primaryClass={cs.LG},
      url={https://arxiv.org/abs/2506.07549}, 
}

@INPROCEEDINGS{horowitz2014energy,
  author={Horowitz, Mark},
  booktitle={2014 IEEE International Solid-State Circuits Conference Digest of Technical Papers (ISSCC)}, 
  title={1.1 Computing's energy problem (and what we can do about it)}, 
  year={2014},
  volume={},
  number={},
  pages={10-14},
  doi={10.1109/ISSCC.2014.6757323}}

@misc{fan2025shapkan,
      title={Shift-Invariant Attribute Scoring for Kolmogorov-Arnold Networks via Shapley Value}, 
      author={Wangxuan Fan and Ching Wang and Siqi Li and Nan Liu},
      year={2026},
      eprint={2510.01663},
      archivePrefix={arXiv},
      primaryClass={cs.LG},
      url={https://arxiv.org/abs/2510.01663}, 
}

@misc{ta2025prkan,
      title={PRKAN: Parameter-Reduced Kolmogorov-Arnold Networks}, 
      author={Hoang-Thang Ta and Duy-Quy Thai and Anh Tran and Grigori Sidorov and Alexander Gelbukh},
      year={2025},
      eprint={2501.07032},
      archivePrefix={arXiv},
      primaryClass={cs.LG},
      url={https://arxiv.org/abs/2501.07032}, 
}

@misc{jacob2018quantization,
      title={Quantization and Training of Neural Networks for Efficient Integer-Arithmetic-Only Inference}, 
      author={Benoit Jacob and Skirmantas Kligys and Bo Chen and Menglong Zhu and Matthew Tang and Andrew Howard and Hartwig Adam and Dmitry Kalenichenko},
      year={2017},
      eprint={1712.05877},
      archivePrefix={arXiv},
      primaryClass={cs.LG},
      url={https://arxiv.org/abs/1712.05877}, 
}

@inproceedings{li2025lipkan,
  author    = {Li, Pengqi and Ding, Lizhong and Fu, Jiarun and Zhang, Chunhui and
             Wang, Guoren and Yuan, Ye},
  title     = {Generalization Bounds for Kolmogorov-Arnold Networks ({KANs}) and
               Enhanced {KANs} with Lower Lipschitz Complexity},
  booktitle = {Advances in Neural Information Processing Systems (NeurIPS)},
  year      = {2025}
}

@inproceedings{beyer1999nn,
  author    = {Beyer, Kevin and Goldstein, Jonathan and Ramakrishnan, Raghu and Shaft, Uri},
  title     = {When Is ``Nearest Neighbor'' Meaningful?},
  booktitle = {International Conference on Database Theory (ICDT)},
  pages     = {217--235},
  publisher = {Springer},
  year      = {1999}
}

@article{donoho2000curses,
  author  = {Donoho, David L.},
  title   = {High-Dimensional Data Analysis: The Curses and Blessings of Dimensionality},
  journal = {AMS Math Challenges Lecture},
  volume  = {1},
  pages   = {1--32},
  year    = {2000}
}

@book{gersho1992vq,
  author    = {Gersho, Allen and Gray, Robert M.},
  title     = {Vector Quantization and Signal Compression},
  publisher = {Kluwer Academic Publishers},
  address   = {Norwell, MA},
  series    = {The Kluwer International Series in Engineering and Computer Science},
  year      = {1992}
}

@article{sze2017efficient,
  author  = {Sze, Vivienne and Chen, Yu-Hsin and Yang, Tien-Ju and Emer, Joel S.},
  title   = {Efficient Processing of Deep Neural Networks: A Tutorial and Survey},
  journal = {Proceedings of the IEEE},
  volume  = {105},
  number  = {12},
  pages   = {2295--2329},
  year    = {2017}
}

\clearpage
\appendix

\section{FuncCode Appendix}

This appendix carries the evidence that is intentionally compressed in the
main paper so that the central narrative remains visible: the full ten-seed
MNIST/CIFAR benchmark, metric and normalization controls, matched-storage and
low-rank baselines, published sharing baselines, robustness, precision and
design sweeps, convolutional-KAGN diagnostics, and the complete FPGA reports.
The ten-seed benchmark (seeds 0--9, width 64, 10 training epochs, 20
codebook-finetuning epochs) is the primary fully connected evidence and
supersedes earlier three-seed tables wherever they overlap. The convolutional
CIFAR track is a separate three-seed protocol with 200 dense-training epochs
and equal 30-epoch fine-tuning for compressed arms.

Accuracy is top-1 test accuracy in percent. For the fully connected track,
``W4'' denotes the hardware-quantized codebook stage
(\texttt{clustered\_hwq\_w4}); the convolutional CIFAR results instead use
w8 codebooks because the w4 ablation is unstable despite negligible storage
savings. Compression is always measured against the same family's dense FP32
model using the bit-exact export. Hardware accuracies are reported under
deployed arithmetic and should not be numerically mixed with training-time
software accuracies.

\section{Full Hardware Detail}

This section reports every resource number behind the hardware summary in the main paper:
both synthesis stages for all nine accelerators, the stage-by-stage
decomposition of the gains, the verification ladder, the exact storage
breakdown, and the protocol attempts that were tried and discarded. The
organizing fact to keep in view is that within each family the two quantized
designs differ only in the weight fetch, so any resource difference between
them is caused by the memory format and by nothing else.

\subsection{All nine designs: post-synthesis}

Table~\ref{tab:hwfull} gives the post-synthesis estimates on the
\texttt{xczu9eg}, the part chosen so that even the largest FP32 baseline fits
on-chip. Read the BRAM18 column down each family: the FP32 baseline is
memory-dominated, the INT4 design cuts that memory by roughly the precision
ratio, and \funccode{} cuts it again without touching latency. The LUT and FF
columns show the other half of the story, namely that the index indirection is
nearly free in logic: \funccode{} uses slightly \emph{fewer} LUTs than its
INT4 sibling in every family, because the small codebooks it substitutes for a
large weight ROM are cheaper to address.

\begin{table*}[!ht]
\centering
\caption{Post-synthesis, Vivado HLS 2019.1, \texttt{xczu9eg-ffvb1156-2-e},
6.67\,ns target. Latency = worst-case cycles $\times$ 6.67\,ns, batch~1.
Within a family the LSQ and \funccode{} designs have identical cycle counts
and II=1 fused main loops.}
\label{tab:hwfull}
\small
\begin{tabular}{lrrrrrr}
\toprule
Design & Clk(ns) & LUT & FF & DSP & BRAM18 & Lat.($\mu$s) \\
\midrule
D1 spline FP32   & 5.78  & 15928 & 15328 & 48  & 842 & 7086.6 \\
D2 spline LSQ    & 6.50  & 3374  & 5457  & 24  & 108 & 340.1 \\
D3 spline \funccode{} & 6.50 & 3197 & 5523 & 23 & \textbf{32} & 340.1 \\
\midrule
G1 GRAM FP32     & 5.78  & 17935 & 15846 & 62  & 469 & 1933.9 \\
G2 GRAM LSQ      & 6.60  & 34571 & 42863 & 52  & 76  & 343.6 \\
G3 GRAM \funccode{} & 6.60 & 34381 & 42893 & 51 & \textbf{43} & 343.6 \\
\midrule
K1 conv FP32     & 11.39$^{\dagger}$ & 37810 & 28421 & 109 & 96 & 12525.3 \\
K2 conv LSQ      & 6.70$^{\ddagger}$ & 61532 & 71169 & 84  & 41 & 1963.1 \\
K3 conv \funccode{} & 6.70$^{\ddagger}$ & 61008 & 70873 & 82 & \textbf{36} & 1963.1 \\
\bottomrule
\end{tabular}
\\[2pt]\raggedright\footnotesize
$^{\dagger}$Synthesis estimation pessimism for the floating-point cores;
post-route timing is met at 5.30\,ns, so the latency quoted at the target
period is valid. $^{\ddagger}$Estimate marginally above target; post-route met
with $\geq$0.7\,ns slack.
\end{table*}

\subsection{All nine designs: post-implementation}

Synthesis estimates are not silicon. Table~\ref{tab:impl} therefore repeats
the measurement after place-and-route on the licensed \texttt{xczu7ev}, where
every design that fits the part meets timing. The BRAM ordering is unchanged
and the \funccode{} advantage in fact grows slightly
($3.4\times \rightarrow 3.87\times$ on spline), which is the outcome one
wants: the effect survives the more pessimistic stage. The one absence is the
spline FP32 baseline, whose FP32 weight ROMs do not fit this part; rather than
substitute a synthesis estimate for it we leave the cell empty, and we never
mix the two stages inside a single comparison.

\begin{table*}[!ht]
\centering
\caption{Post-implementation (place-and-route), \texttt{xczu7ev-ffvc1156-2-e},
6.67\,ns constraint. Achieved period = constraint $-$ WNS. All designs that
fit the part met timing.}
\label{tab:impl}
\small
\begin{tabular}{lrrrrrl}
\toprule
Design & Clk(ns) & LUT & FF & DSP & BRAM18 & Timing \\
\midrule
D1 spline FP32   & --   & --    & --    & --  & --  & n/a$^{*}$ \\
D2 spline LSQ    & 6.56 & 3546  & 4935  & 32  & 147 & met \\
D3 spline \funccode{} & 6.20 & 3554 & 4936 & 31 & \textbf{38} & met \\
\midrule
G1 GRAM FP32     & 6.22 & 15084 & 13393 & 62  & 608 & met \\
G2 GRAM LSQ      & 6.27 & 32019 & 39345 & 64  & 90  & met \\
G3 GRAM \funccode{} & 6.20 & 31934 & 39346 & 59 & \textbf{46} & met \\
\midrule
K1 conv FP32     & \textbf{5.30} & 26870 & 22968 & 109 & 118 & met$^{\S}$ \\
K2 conv LSQ      & 5.93 & 52634 & 64956 & 120 & 38  & met \\
K3 conv \funccode{} & 5.89 & 52966 & 64892 & 120 & \textbf{32} & met \\
\bottomrule
\end{tabular}
\\[2pt]\raggedright\footnotesize
$^{*}$The spline FP32 baseline's 397 BRAM36 of FP32 weight ROMs exceed this
part. Post-synthesis and post-implementation numbers are never mixed in any
comparison. $^{\S}$$+1.371$\,ns slack, 0/45{,}480 endpoints failing.
\end{table*}

\subsection{Stage-by-stage gains}

The decomposition of the total win into its two causes (FP32$\to$INT4, a
precision change; INT4$\to$\funccode{}, a memory-format change) is Table~6 of
the main paper. The key entry is the latency row of the second block: it is
$1.00\times$ in all three families, to the cycle, which is what makes the BRAM
saving causally attributable to the weight format.

\subsection{Verification ladder}

All nine designs completed every rung.
\textbf{L1} (PyTorch deployed-arithmetic vs.\ numpy emulator, all 10{,}000
MNIST test images): argmax 10000/10000 for all six quantized designs; the
three FP32 baselines match within $10^{-4}$--$10^{-3}$ with exact argmax.
\textbf{L2} (HLS C simulation vs.\ emulator INT32 logits, 1{,}000 stratified
golden vectors): \emph{bit-exact} 1000/1000 for all six quantized designs.
\textbf{L3} (RTL co-simulation vs.\ C simulation, 50 vectors): \emph{bit-exact}
50/50 for all six quantized designs; FP32 baselines pass 50/50 within float
tolerance. \textbf{L4} (deployed accuracy vs.\ FP32 dense reference): the
per-design deltas are in the stage-latency table of the main paper.
Golden vectors are 1{,}000 stratified test images (first 100 per class,
deterministic, SHA256-pinned); co-simulation uses the first 50, identical for
every design.

\subsection{Storage breakdown of the deployed models}

Table~\ref{tab:store} is the analytical prediction that the hardware was built
to test. The index-share column is the claim: once codebooks are quantized to
four bits, they occupy a few hundred bytes and essentially all remaining
storage is pointers. The fully connected families sit at $98.9\%$ and
$99.1\%$, which is as index-bound as a model can be. The convolutional family
is the informative exception at $82.3\%$: because a conv KAN already shares its
edge functions across spatial positions, it has far fewer edges relative to its
codebooks, so the index stream does not dominate as completely, and this is
precisely the family in which \funccode{} buys the least hardware.

\begin{table}[h]
\centering
\caption{Bit-exact analytical storage. The index-stream share is what
Eq.~(3) of the main paper predicts and what the BRAM measurements confirm.}
\label{tab:store}
\footnotesize
\setlength{\tabcolsep}{2.5pt}
\begin{tabular}{lrrr}
\toprule
Design & Weight storage & Compression & Index share \\
\midrule
D1 spline FP32       & 1786.500\,KiB & 1.00$\times$ & -- \\
D2 spline LSQ        & 223.336\,KiB  & 8.00$\times$ & -- \\
D3 spline \funccode{}& 56.469\,KiB   & \textbf{31.64}$\times$ & \textbf{98.9\%} \\
G1 GRAM FP32         & 992.500\,KiB  & 1.00$\times$ & -- \\
G2 GRAM LSQ          & 124.086\,KiB  & 8.00$\times$ & -- \\
G3 GRAM \funccode{}  & 56.344\,KiB   & \textbf{17.62}$\times$ & \textbf{99.1\%} \\
K1 conv FP32         & 99.063\,KiB   & 1.00$\times$ & -- \\
K2 conv LSQ          & 12.422\,KiB   & 7.98$\times$ & -- \\
K3 conv \funccode{}  & 7.527\,KiB    & \textbf{13.16}$\times$ & \textbf{82.3\%} \\
\bottomrule
\end{tabular}
\end{table}

\subsection{Protocol exploration (recorded, superseded)}

We record the attempts that did not become the final protocol, since they are
the evidence behind our design choices.
Post-hoc W4 PTQ of a float-trained conv KAGN reaches only \textbf{23.4\%}
(about a quarter of conv weights round to zero), which motivated
from-scratch QAT for the convolutional family. Initializing the conv INT4
design from the dense checkpoint gives 91.22\% against 92.74\% from scratch.
Clustering the conv \funccode{} design from the FP32 dense model gives
66.44\%, against 89.78\% when clustered from the quantized model with
$K_s{=}64$. For the spline \funccode{} design, a 10-epoch codebook finetune
gives 95.07\% and a 50-epoch one 95.16\%, so the 25-epoch schedule (95.22\%
deployed) is kept.

\section{Full Ten-Seed Benchmark}
\label{apx:bench}
Tables~\ref{tab:apx-mnist}--\ref{tab:apx-c100} report every arm of the primary
benchmark at both the finetuned FP32-codebook stage and the deployed W4 stage.
The pre-finetune (zero-shot) stage for every configuration is in the released
CSVs.

\begin{table}[!ht]
\centering
\caption{Ten-seed benchmark, MNIST. FP32-FT is the finetuned FP32-codebook stage; W4 is the deployed hardware-quantized stage. Compression against the family's dense model.}
\label{tab:apx-mnist}
\footnotesize
\setlength{\tabcolsep}{3pt}
\begin{tabular}{llccr}
\toprule
Family & Method & FP32-FT & W4 & Comp. \\
\midrule
spline & dense & 95.86 $\pm$ 0.20 & -- & 1.0$\times$ \\
spline & coeff.\ $K{=}16$ & 92.48 $\pm$ 0.44 & 90.19 $\pm$ 1.97 & 71.2$\times$ \\
spline & coeff.\ $K{=}32$ & 93.99 $\pm$ 0.31 & 93.22 $\pm$ 0.53 & 56.6$\times$ \\
spline & func.\ $K{=}16$ & 92.92 $\pm$ 0.35 & 92.48 $\pm$ 0.43 & 71.2$\times$ \\
spline & func.\ $K{=}32$ & 93.91 $\pm$ 0.37 & 92.96 $\pm$ 1.47 & 56.6$\times$ \\
spline & branch $(16,8)$ & 95.22 $\pm$ 0.30 & 95.00 $\pm$ 0.52 & 40.8$\times$ \\
spline & branch $(32,16)$ & 95.68 $\pm$ 0.19 & 95.55 $\pm$ 0.31 & 31.6$\times$ \\
\midrule
FastKAN & dense & 96.83 $\pm$ 0.22 & -- & 1.0$\times$ \\
FastKAN & coeff.\ $K{=}16$ & 95.09 $\pm$ 0.23 & 94.25 $\pm$ 0.82 & 71.2$\times$ \\
FastKAN & coeff.\ $K{=}32$ & 95.75 $\pm$ 0.20 & 95.42 $\pm$ 0.38 & 56.6$\times$ \\
FastKAN & func.\ $K{=}16$ & 92.62 $\pm$ 0.36 & 78.12 $\pm$ 11.79 & 71.2$\times$ \\
FastKAN & func.\ $K{=}32$ & 93.59 $\pm$ 0.31 & 86.83 $\pm$ 11.70 & 56.6$\times$ \\
FastKAN & branch $(16,8)$ & 94.19 $\pm$ 0.16 & 93.13 $\pm$ 1.22 & 40.8$\times$ \\
FastKAN & branch $(32,16)$ & 94.56 $\pm$ 0.21 & 93.85 $\pm$ 1.26 & 31.6$\times$ \\
\midrule
GRAM & dense & 96.74 $\pm$ 0.08 & -- & 1.0$\times$ \\
GRAM & coeff.\ $K{=}16$ & 95.20 $\pm$ 0.32 & 94.91 $\pm$ 0.34 & 39.7$\times$ \\
GRAM & coeff.\ $K{=}32$ & 95.90 $\pm$ 0.20 & 95.60 $\pm$ 0.27 & 31.6$\times$ \\
GRAM & func.\ $K{=}16$ & 94.89 $\pm$ 0.41 & 94.62 $\pm$ 0.66 & 39.7$\times$ \\
GRAM & func.\ $K{=}32$ & 95.68 $\pm$ 0.20 & 95.46 $\pm$ 0.27 & 31.6$\times$ \\
GRAM & branch $(16,8)$ & 96.11 $\pm$ 0.19 & 95.79 $\pm$ 0.21 & 22.7$\times$ \\
GRAM & branch $(32,16)$ & 96.58 $\pm$ 0.13 & 96.40 $\pm$ 0.21 & 17.6$\times$ \\
\midrule
MLP & dense & 96.06 $\pm$ 0.19 & -- & 1.0$\times$ \\
MLP & uniform W4 PTQ & -- & 95.66 $\pm$ 0.24 & 8.0$\times$ \\
MLP & LSQ QAT W4 & -- & 97.36 $\pm$ 0.09 & 8.0$\times$ \\
\bottomrule
\end{tabular}
\end{table}

\begin{table}[!ht]
\centering
\caption{Ten-seed benchmark, CIFAR-10. FP32-FT is the finetuned FP32-codebook stage; W4 is the deployed hardware-quantized stage. Compression against the family's dense model.}
\label{tab:apx-c10}
\footnotesize
\setlength{\tabcolsep}{3pt}
\begin{tabular}{llccr}
\toprule
Family & Method & FP32-FT & W4 & Comp. \\
\midrule
spline & dense & 46.82 $\pm$ 0.44 & -- & 1.0$\times$ \\
spline & coeff.\ $K{=}16$ & 37.47 $\pm$ 1.67 & 28.88 $\pm$ 6.54 & 71.8$\times$ \\
spline & coeff.\ $K{=}32$ & 41.64 $\pm$ 0.96 & 34.52 $\pm$ 3.85 & 57.3$\times$ \\
spline & func.\ $K{=}16$ & 36.41 $\pm$ 1.16 & 30.71 $\pm$ 3.63 & 71.8$\times$ \\
spline & func.\ $K{=}32$ & 41.13 $\pm$ 0.67 & 36.80 $\pm$ 2.57 & 57.3$\times$ \\
spline & branch $(16,8)$ & 42.46 $\pm$ 0.44 & 38.10 $\pm$ 3.31 & 41.1$\times$ \\
spline & branch $(32,16)$ & 44.91 $\pm$ 0.40 & 42.03 $\pm$ 1.59 & 31.9$\times$ \\
\midrule
FastKAN & dense & 50.04 $\pm$ 0.60 & -- & 1.0$\times$ \\
FastKAN & coeff.\ $K{=}16$ & 44.72 $\pm$ 0.46 & 42.32 $\pm$ 2.68 & 71.8$\times$ \\
FastKAN & coeff.\ $K{=}32$ & 46.51 $\pm$ 0.75 & 45.39 $\pm$ 1.43 & 57.3$\times$ \\
FastKAN & func.\ $K{=}16$ & 36.07 $\pm$ 0.67 & 34.10 $\pm$ 4.48 & 71.8$\times$ \\
FastKAN & func.\ $K{=}32$ & 38.97 $\pm$ 0.61 & 36.77 $\pm$ 3.64 & 57.3$\times$ \\
FastKAN & branch $(16,8)$ & 37.45 $\pm$ 0.85 & 35.73 $\pm$ 1.86 & 41.1$\times$ \\
FastKAN & branch $(32,16)$ & 37.46 $\pm$ 1.14 & 36.26 $\pm$ 2.02 & 31.9$\times$ \\
\midrule
GRAM & dense & 47.17 $\pm$ 0.44 & -- & 1.0$\times$ \\
GRAM & coeff.\ $K{=}16$ & 45.74 $\pm$ 0.58 & 40.64 $\pm$ 4.38 & 39.9$\times$ \\
GRAM & coeff.\ $K{=}32$ & 47.28 $\pm$ 0.68 & 45.19 $\pm$ 1.97 & 31.9$\times$ \\
GRAM & func.\ $K{=}16$ & 45.33 $\pm$ 0.61 & 42.31 $\pm$ 2.94 & 39.9$\times$ \\
GRAM & func.\ $K{=}32$ & 46.34 $\pm$ 0.74 & 44.80 $\pm$ 1.68 & 31.9$\times$ \\
GRAM & branch $(16,8)$ & 46.55 $\pm$ 0.73 & 42.20 $\pm$ 4.08 & 22.8$\times$ \\
GRAM & branch $(32,16)$ & 47.46 $\pm$ 0.66 & 44.82 $\pm$ 2.55 & 17.7$\times$ \\
\midrule
MLP & dense & 43.93 $\pm$ 0.82 & -- & 1.0$\times$ \\
MLP & uniform W4 PTQ & -- & 42.49 $\pm$ 1.24 & 8.0$\times$ \\
MLP & LSQ QAT W4 & -- & 48.47 $\pm$ 0.55 & 8.0$\times$ \\
\bottomrule
\end{tabular}
\end{table}

\begin{table}[!ht]
\centering
\caption{Ten-seed benchmark, CIFAR-100. FP32-FT is the finetuned FP32-codebook stage; W4 is the deployed hardware-quantized stage. Compression against the family's dense model.}
\label{tab:apx-c100}
\footnotesize
\setlength{\tabcolsep}{3pt}
\begin{tabular}{llccr}
\toprule
Family & Method & FP32-FT & W4 & Comp. \\
\midrule
spline & dense & 14.21 $\pm$ 3.15 & -- & 1.0$\times$ \\
spline & coeff.\ $K{=}16$ & 5.92 $\pm$ 1.53 & 4.35 $\pm$ 1.34 & 71.8$\times$ \\
spline & coeff.\ $K{=}32$ & 6.67 $\pm$ 1.73 & 5.15 $\pm$ 1.63 & 57.4$\times$ \\
spline & func.\ $K{=}16$ & 7.16 $\pm$ 1.79 & 6.03 $\pm$ 1.93 & 71.8$\times$ \\
spline & func.\ $K{=}32$ & 8.42 $\pm$ 2.11 & 7.56 $\pm$ 2.33 & 57.4$\times$ \\
spline & branch $(16,8)$ & 10.79 $\pm$ 2.44 & 9.87 $\pm$ 2.31 & 41.1$\times$ \\
spline & branch $(32,16)$ & 13.24 $\pm$ 3.03 & 12.27 $\pm$ 2.93 & 31.9$\times$ \\
\midrule
FastKAN & dense & 22.10 $\pm$ 0.35 & -- & 1.0$\times$ \\
FastKAN & coeff.\ $K{=}16$ & 15.35 $\pm$ 0.55 & 14.75 $\pm$ 0.63 & 71.8$\times$ \\
FastKAN & coeff.\ $K{=}32$ & 16.92 $\pm$ 0.57 & 15.99 $\pm$ 0.90 & 57.4$\times$ \\
FastKAN & func.\ $K{=}16$ & 6.48 $\pm$ 0.66 & 4.62 $\pm$ 1.82 & 71.8$\times$ \\
FastKAN & func.\ $K{=}32$ & 8.65 $\pm$ 0.80 & 6.90 $\pm$ 1.53 & 57.4$\times$ \\
FastKAN & branch $(16,8)$ & 7.80 $\pm$ 0.90 & 7.17 $\pm$ 1.37 & 41.1$\times$ \\
FastKAN & branch $(32,16)$ & 8.04 $\pm$ 1.33 & 7.22 $\pm$ 1.21 & 31.9$\times$ \\
\midrule
GRAM & dense & 20.94 $\pm$ 0.44 & -- & 1.0$\times$ \\
GRAM & coeff.\ $K{=}16$ & 18.90 $\pm$ 0.58 & 17.09 $\pm$ 1.64 & 39.9$\times$ \\
GRAM & coeff.\ $K{=}32$ & 19.87 $\pm$ 0.35 & 18.21 $\pm$ 0.64 & 31.9$\times$ \\
GRAM & func.\ $K{=}16$ & 16.97 $\pm$ 0.57 & 15.83 $\pm$ 1.08 & 39.9$\times$ \\
GRAM & func.\ $K{=}32$ & 18.57 $\pm$ 0.57 & 17.54 $\pm$ 1.20 & 31.9$\times$ \\
GRAM & branch $(16,8)$ & 19.32 $\pm$ 0.40 & 17.69 $\pm$ 1.83 & 22.8$\times$ \\
GRAM & branch $(32,16)$ & 20.29 $\pm$ 0.32 & 19.48 $\pm$ 0.49 & 17.7$\times$ \\
\midrule
MLP & dense & 17.35 $\pm$ 0.34 & -- & 1.0$\times$ \\
MLP & uniform W4 PTQ & -- & 14.92 $\pm$ 0.55 & 8.0$\times$ \\
MLP & LSQ QAT W4 & -- & 20.55 $\pm$ 0.59 & 8.0$\times$ \\
\bottomrule
\end{tabular}
\end{table}

\section{Normalization and Metric Controls}
\label{apx:controls}
The control family re-runs single-codebook clustering at $K{=}32$ with five
seeds while varying only the preprocessing: coefficient clustering raw,
standardized, or whitened, and function or branch clustering with signature
normalization disabled. Table~\ref{tab:apx-ctrl} reports the deployed W4
stage.

\begin{table}[!ht]
\centering
\caption{Normalization controls at $K{=}32$, W4, five seeds. \emph{unnorm}
disables signature normalization; \emph{std}/\emph{whiten} standardize or
whiten coefficients before clustering.}
\label{tab:apx-ctrl}
\footnotesize
\setlength{\tabcolsep}{3pt}
\begin{tabular}{llcc}
\toprule
Family & Variant & MNIST & CIFAR-10 \\
\midrule
spline & coefficient        & 93.35 $\pm$ 0.64 & 34.76 $\pm$ 2.06 \\
spline & coefficient\_std    & 91.26 $\pm$ 0.83 & 35.31 $\pm$ 1.89 \\
spline & coefficient\_whiten & 88.68 $\pm$ 1.60 & 31.60 $\pm$ 3.97 \\
spline & function            & 93.21 $\pm$ 0.52 & 36.57 $\pm$ 3.34 \\
spline & function\_unnorm    & 95.49 $\pm$ 0.18 & 37.13 $\pm$ 1.65 \\
spline & branch $(32,16)$    & 95.49 $\pm$ 0.25 & 40.51 $\pm$ 4.63 \\
spline & branch\_unnorm      & 95.66 $\pm$ 0.15 & 41.19 $\pm$ 1.76 \\
\midrule
FastKAN & coefficient        & 95.09 $\pm$ 0.52 & 43.62 $\pm$ 3.37 \\
FastKAN & coefficient\_std    & 95.19 $\pm$ 0.37 & 45.85 $\pm$ 0.62 \\
FastKAN & coefficient\_whiten & 93.81 $\pm$ 0.35 & 44.71 $\pm$ 0.36 \\
FastKAN & function            & 95.55 $\pm$ 0.33 & 44.55 $\pm$ 1.21 \\
FastKAN & function\_unnorm    & 96.20 $\pm$ 0.29 & 46.36 $\pm$ 0.75 \\
FastKAN & branch $(32,16)$    & 96.39 $\pm$ 0.20 & 47.86 $\pm$ 0.76 \\
FastKAN & branch\_unnorm      & 96.40 $\pm$ 0.12 & 48.95 $\pm$ 0.43 \\
\midrule
GRAM & coefficient        & 95.33 $\pm$ 0.20 & 45.33 $\pm$ 0.67 \\
GRAM & coefficient\_std    & 95.32 $\pm$ 0.27 & 44.41 $\pm$ 2.58 \\
GRAM & coefficient\_whiten & 94.93 $\pm$ 0.39 & 43.07 $\pm$ 3.39 \\
GRAM & function            & 95.33 $\pm$ 0.21 & 44.79 $\pm$ 1.28 \\
GRAM & function\_unnorm    & 95.76 $\pm$ 0.18 & 44.46 $\pm$ 0.72 \\
GRAM & branch $(32,16)$    & 96.28 $\pm$ 0.22 & 44.80 $\pm$ 2.29 \\
GRAM & branch\_unnorm      & 96.09 $\pm$ 0.23 & 44.19 $\pm$ 2.47 \\
\bottomrule
\end{tabular}
\end{table}

The six paired function-minus-coefficient differences at $K{=}32$/W4 are
$-0.14 \pm 0.92$ (spline, MNIST), $-0.00 \pm 0.37$ (GRAM, MNIST),
$+0.46 \pm 0.52$ (FastKAN, MNIST), $+1.81 \pm 2.69$ (spline, CIFAR-10),
$-0.54 \pm 1.88$ (GRAM, CIFAR-10) and $+0.93 \pm 2.95$ (FastKAN, CIFAR-10);
every interval includes zero. Function against function\_unnorm on
GRAM/CIFAR-10 is $+0.33 \pm 1.96$ and branch against branch\_unnorm is
$+0.61 \pm 3.18$, so signature normalization is cosmetic at this operating
point.

\section{Low-Rank Factorization Baseline}
\label{apx:lowrank}

\begin{table}[!ht]
\centering
\caption{Low-rank baseline, five seeds. FP32-FT is the finetuned
factorization; W8 and W4 quantize the factors. Compression shown for W4.}
\label{tab:apx-lowrank}
\footnotesize
\setlength{\tabcolsep}{2.6pt}
\begin{tabular}{llcccr}
\toprule
 & $r$ & FP32-FT & W8 & W4 & Comp.\ (W4) \\
\midrule
\multicolumn{6}{l}{\emph{MNIST, spline}}\\
 & 1 & 96.19 $\pm$ 0.30 & 96.19 $\pm$ 0.28 & 95.36 $\pm$ 0.33 & 71.9$\times$ \\
 & 2 & 96.61 $\pm$ 0.28 & 96.60 $\pm$ 0.25 & 95.60 $\pm$ 0.69 & 36.0$\times$ \\
 & 3 & 96.72 $\pm$ 0.05 & 96.72 $\pm$ 0.08 & 94.02 $\pm$ 1.42 & 24.0$\times$ \\
 & 4 & 96.65 $\pm$ 0.23 & 96.65 $\pm$ 0.24 & 94.50 $\pm$ 1.32 & 18.0$\times$ \\
\multicolumn{6}{l}{\emph{MNIST, GRAM}}\\
 & 1 & 96.91 $\pm$ 0.08 & 96.90 $\pm$ 0.12 & 95.18 $\pm$ 1.20 & 40.0$\times$ \\
 & 2 & 97.00 $\pm$ 0.15 & 97.01 $\pm$ 0.14 & 90.70 $\pm$ 7.02 & 20.0$\times$ \\
 & 3 & 97.03 $\pm$ 0.10 & 97.01 $\pm$ 0.09 & 92.14 $\pm$ 1.59 & 13.3$\times$ \\
 & 4 & 97.09 $\pm$ 0.12 & 97.09 $\pm$ 0.15 & 89.86 $\pm$ 5.11 & 10.0$\times$ \\
\multicolumn{6}{l}{\emph{CIFAR-10, spline}}\\
 & 1 & 46.76 $\pm$ 1.43 & 46.78 $\pm$ 1.46 & 45.16 $\pm$ 2.20 & 72.0$\times$ \\
 & 2 & 51.09 $\pm$ 0.51 & 51.09 $\pm$ 0.42 & 44.44 $\pm$ 4.82 & 36.0$\times$ \\
 & 3 & 51.30 $\pm$ 0.22 & 51.28 $\pm$ 0.15 & 40.47 $\pm$ 4.33 & 24.0$\times$ \\
 & 4 & 49.87 $\pm$ 0.41 & 49.76 $\pm$ 0.35 & 31.44 $\pm$ 4.80 & 18.0$\times$ \\
\multicolumn{6}{l}{\emph{CIFAR-10, GRAM}}\\
 & 1 & 48.12 $\pm$ 0.54 & 48.14 $\pm$ 0.48 & 44.68 $\pm$ 2.35 & 40.0$\times$ \\
 & 2 & 50.54 $\pm$ 0.70 & 50.48 $\pm$ 0.62 & 47.74 $\pm$ 1.04 & 20.0$\times$ \\
 & 3 & 50.92 $\pm$ 0.36 & 50.78 $\pm$ 0.39 & 37.25 $\pm$ 7.03 & 13.3$\times$ \\
 & 4 & 50.65 $\pm$ 0.35 & 50.60 $\pm$ 0.43 & 34.31 $\pm$ 4.41 & 10.0$\times$ \\
\bottomrule
\end{tabular}
\end{table}

At W8 the factorization is essentially lossless and very strong; at W4 it
becomes non-monotone in rank, with ranks 3 and 4 collapsing on CIFAR-10 and
GRAM rank 2 unstable on MNIST. The main paper reports the honest comparison at
the deployed W4 point.

\section{LSQ-Quantized MLP}
\label{apx:lsq}
The LSQ QAT baseline (five seeds, 30 QAT epochs) reaches
$97.36 \pm 0.09$ on MNIST at $24.8203$\,KiB, $48.47 \pm 0.55$ on CIFAR-10 at
$96.3203$\,KiB and $20.55 \pm 0.59$ on CIFAR-100 at $99.1328$\,KiB. Its budget
exceeds the KAN protocol (10 training plus 20 finetune epochs) and on MNIST it
surpasses its own dense reference ($96.06 \pm 0.19$), so the numbers should be
read as a strong external yardstick rather than a matched comparison.

\section{Codebook Precision Under Stress: PTQ Against QAT}
\label{apx:stress}
On CIFAR-10 at $(32,16)$, post-training quantization of the codebooks and
quantization-aware finetuning agree at moderate precision and separate at low
precision. At 4-bit codebooks GRAM reads $44.73 \pm 2.75$ (PTQ) against
$46.86 \pm 0.85$ (QAT) and spline $40.81 \pm 1.00$ against $44.59 \pm 0.95$;
at 2-bit the gap widens to $26.43 \pm 7.45$ against $41.85 \pm 2.30$ (GRAM)
and $20.11 \pm 8.44$ against $35.81 \pm 1.74$ (spline). QAT buys back most of
the low-bit loss without touching the index streams.

The grouped-shared-basis (GSB) alternative removes cross-edge sharing and
sweeps the grouped-basis precision instead. Across all 144 GSB configurations
only four reach $\le 10$ bits per edge, with accuracies of $10.2$, $20.4$,
$33.7$ and $39.5$; \funccode{} operates at nine bits per edge without
collapse. Full sweeps, including the separate-base variant, are in the
released CSVs.

\section{MetaCluster Baseline}
\label{apx:meta}
The MetaCluster reimplementation passes its fidelity gate for FastKAN
($48.21 \pm 1.08$ against an expected $47.6$) and GRAM ($48.79 \pm 0.27$
against $48.47$), and at width 64 the GRAM arm reaches $48.38 \pm 0.71$
($K{=}16$) and $48.70 \pm 0.85$ ($K{=}32$) after finetuning. The spline arm
collapsed in a seed-dependent way ($24.14 \pm 19.38$ at the anchor against an
expected $46.28$, propagating to $18.34 \pm 17.58$ at width 64) and fails the
$\pm 1.5$\,pp gate; per the baselines plan it requires a learning-rate and
embedding-initialization retune before it can be reported, so we exclude it
rather than publish a broken arm. Under codebook quantization the GRAM arm degrades sharply below eight
bits: at $K{=}32$ the PTQ sweep reads $47.55 \pm 0.41$ (cb8), $33.64 \pm
6.55$ (cb6), $17.60 \pm 10.08$ (cb4) and $10.76 \pm 2.25$ (cb2), with QAT
recovering only partially ($48.45 \pm 0.81$, $44.51 \pm 2.84$, $30.76 \pm
6.38$, $12.56 \pm 1.64$); the $K{=}16$ family behaves the same. The FastKAN
anchor is robust across the whole sweep ($48.03 \pm 1.07$ at QAT cb4,
$46.74 \pm 1.96$ at PTQ cb4), so the quantization fragility is
family-dependent.

\section{Soft-Branch Assignment, Multi-Seed}
\label{apx:soft}
Across four three-seed configurations (two codebook sizes, softmax and
straight-through estimators), the learned assignment finishes below the hard
$k$-means baseline at the deployed W4 stage in every case:
$-0.75 \pm 1.53$, $-1.22 \pm 0.71$, $-0.76 \pm 1.56$ and $-0.39 \pm 0.46$\,pp
(learned minus hard). The single-seed probe in the main paper is therefore
conservative.

\section{Out-of-Distribution Robustness}
\label{apx:ood}

\begin{table}[!ht]
\centering
\caption{MNIST accuracy under input corruption (seed 42, $K{=}32$ arms).}
\label{tab:apx-ood}
\footnotesize
\setlength{\tabcolsep}{3pt}
\begin{tabular}{lcccc}
\toprule
Perturbation & dense & branch & function & coeff. \\
\midrule
gaussian 0.0  & 95.65 & 95.42 & 93.12 & 94.11 \\
gaussian 0.05 & 95.46 & 95.20 & 93.13 & 94.11 \\
gaussian 0.1  & 95.50 & 95.29 & 92.99 & 94.10 \\
gaussian 0.2  & 94.83 & 94.66 & 91.67 & 91.33 \\
gaussian 0.3  & 93.40 & 92.49 & 86.37 & 81.91 \\
gaussian 0.4  & 87.92 & 86.41 & 78.23 & 67.79 \\
\midrule
dropout 0.1   & 94.13 & 93.56 & 88.62 & 88.37 \\
dropout 0.2   & 87.57 & 84.87 & 70.73 & 67.91 \\
dropout 0.3   & 69.95 & 65.80 & 49.42 & 47.52 \\
\midrule
contrast 0.8  & 93.72 & 90.96 & 81.04 & 83.98 \\
contrast 0.6  & 44.74 & 30.68 & 22.27 & 23.53 \\
contrast 0.4  & 10.94 & 10.40 & 10.31 & 10.83 \\
\bottomrule
\end{tabular}
\end{table}

\section{Index Entropy and Huffman Headroom}
\label{apx:entropy}

\begin{table}[!ht]
\centering
\caption{Index-stream coding headroom per checkpoint (bits).}
\label{tab:apx-entropy}
\footnotesize
\setlength{\tabcolsep}{2.6pt}
\begin{tabular}{lrrr}
\toprule
Checkpoint & Fixed & Entropy & Huffman \\
\midrule
MNIST spline branch $K{=}32$    & 462592  & 444529  & 448299 \\
MNIST spline function $K{=}32$  & 258432  & 246753  & 248905 \\
CIFAR-10 GRAM branch $K{=}32$   & 1779456 & 1671502 & 1687837 \\
CIFAR-100 GRAM branch $K{=}32$  & 1831296 & 1712725 & 1727984 \\
\bottomrule
\end{tabular}
\end{table}

In total-storage terms the Huffman savings are $56.5 \to 54.7$\,KiB (3.1\%),
$31.5 \to 30.4$ (3.7\%), $217.2 \to 206.0$ (5.1\%) and $223.5 \to 210.9$
(5.6\%): the fixed-width representation is within a few percent of the entropy
bound.

\section{Effective Rank: Bypass and Initialization}
\label{apx:rank}

\begin{table}[!ht]
\centering
\caption{Function-to-coefficient effective-rank ratio, trained checkpoints,
normalized signatures, five-seed means. FastKAN is shown with and without the
LayerNorm bypass.}
\label{tab:apx-rank}
\footnotesize
\setlength{\tabcolsep}{3.2pt}
\begin{tabular}{llcc}
\toprule
 & Layer & MNIST & CIFAR-10 \\
\midrule
FastKAN (no bypass) & 0 & 0.1124 & 0.1114 \\
FastKAN (no bypass) & 1 & 0.1227 & 0.1147 \\
FastKAN (bypass)    & 0 & 0.7890 & 0.8018 \\
FastKAN (bypass)    & 1 & 0.7655 & 0.7961 \\
spline              & 0 & 0.8506 & 0.8621 \\
spline              & 1 & 0.8187 & 0.8573 \\
GRAM                & 0 & 0.6498 & 0.6640 \\
GRAM                & 1 & 0.6863 & 0.6608 \\
\bottomrule
\end{tabular}
\end{table}

The bypass lifts FastKAN's ratio from about $0.118$ to $0.777$, confirming
the near-one artifact in the main paper is a LayerNorm effect. The
trained-minus-init excess is $-0.0103$ for spline and $-0.0092$ for GRAM
(layer 0, normalized, negative in five of five seeds): the rank gap is present
at initialization and training does not enlarge it, so it is a property of the
basis rather than of the task.

\section{Matched-Storage Small Models}
\label{apx:matched}
The matched-storage comparison in the main paper uses the strongest of three
deliberately small dense configurations (three seeds): a width-16 spline at
grid 5, order 3 reaches $93.77 \pm 0.30$ dense and $92.25 \pm 0.74$ after W4
PTQ at $55.8359$\,KiB; a width-32 spline at grid 1, order 1 reaches
$94.03 \pm 0.14$ after W4 PTQ but sits well below the budget at
$37.2266$\,KiB; a depth-1 GRAM on CIFAR-10 reaches $39.97 \pm 2.06$ after W4 PTQ.
\funccode{} at the same $56$\,KiB reads $95.55 \pm 0.31$.

\section{Single-Seed Design Sweeps}
\label{apx:sweeps}
The single-seed (seed 42) sweeps behind the design-ablation table of the main
paper include a $K$ sweep for both single-codebook metrics at W4:

\begin{table}[!ht]
\centering
\caption{Single-seed $K$ sweep, MNIST spline, W4.}
\label{tab:apx-ksweep}
\footnotesize
\begin{tabular}{lcccc}
\toprule
 & $K{=}16$ & $K{=}32$ & $K{=}64$ & $K{=}128$ \\
\midrule
function    & 92.63 & 93.68 & 94.37 & 95.16 \\
coefficient & 91.87 & 93.05 & 94.12 & 94.85 \\
\bottomrule
\end{tabular}
\end{table}

At seed 42 the function rows lead by $0.76$, $0.63$, $0.25$ and $0.31$\,pp.
The five-seed controls of Section~\ref{apx:controls} show these margins are
within seed noise; we keep the sweep here as the provenance of the earlier
single-seed reading.

\section{Legacy Three-Seed Datasets}
\label{apx:legacy}
The following tables predate the ten-seed protocol (three seeds, earlier
training schedule) and are retained for completeness; where they overlap with
the ten-seed benchmark, the benchmark supersedes them.

\begin{table}[h]
\centering
\caption{Fashion-MNIST, W4, 3 seeds. Spline is the strongest family here,
losing only $0.17$\,pp at $31.64\times$. The GRAM rows carry a warning: the
GRAM \emph{dense} baseline collapses on seed 2026 (per-seed 87.31 / 87.07 /
77.77), which is why its compressed models appear to match or exceed it. That
is a broken baseline, not a gain, and we make no claim from those rows.}
\label{tab:fashion}
\small
\begin{tabular}{llrr}
\toprule
Family & Method & Acc.\ (\%) & Comp. \\
\midrule
spline & dense FP32       & 86.91 $\pm$ 0.38 & 1.00$\times$ \\
spline & branch $(32,16)$ & \textbf{86.73 $\pm$ 0.19} & 31.64$\times$ \\
spline & function $(32)$  & 85.47 $\pm$ 0.21 & 56.63$\times$ \\
\midrule
FastKAN & dense FP32       & 88.43 $\pm$ 0.30 & 1.00$\times$ \\
FastKAN & branch $(32,16)$ & 82.18 $\pm$ 1.10 & 31.64$\times$ \\
FastKAN & function $(32)$  & 83.34 $\pm$ 2.19 & 56.63$\times$ \\
\midrule
GRAM   & dense FP32       & 84.05 $\pm$ 5.44$^{*}$ & 1.00$\times$ \\
GRAM   & branch $(32,16)$ & 84.22 $\pm$ 5.41$^{*}$ & 17.62$\times$ \\
GRAM   & function $(32)$  & 83.77 $\pm$ 5.28$^{*}$ & 31.59$\times$ \\
\midrule
MLP    & dense FP32       & 86.44 $\pm$ 0.22 & 1.00$\times$ \\
MLP    & uniform W4 PTQ   & 84.88 $\pm$ 0.42 & 8.00$\times$ \\
\bottomrule
\end{tabular}
\\[2pt]\raggedright\footnotesize
$^{*}$Unstable dense baseline; see caption.
\end{table}

\begin{table}[h]
\centering
\caption{Tiny ImageNet, W4, single seed (42). Absolute accuracies are low
because the models are fully connected on flattened images, with no
convolutional structure; the comparison between families is still
informative, and it reproduces the CIFAR ordering. GRAM again degrades
gracefully ($-0.62$\,pp) while FastKAN loses most of its accuracy. Spline's
dense model never learns the task ($3.20\%$), so its small compression gap is
not evidence of anything.}
\label{tab:tiny}
\small
\begin{tabular}{llrr}
\toprule
Family & Method & Acc.\ (\%) & Comp. \\
\midrule
spline & dense FP32       & 3.20  & 1.00$\times$ \\
spline & branch $(32,16)$ & 3.09  & 31.99$\times$ \\
\midrule
FastKAN & dense FP32       & 10.45 & 1.00$\times$ \\
FastKAN & branch $(32,16)$ & 2.54  & 31.99$\times$ \\
\midrule
GRAM   & dense FP32       & 11.08 & 1.00$\times$ \\
GRAM   & branch $(32,16)$ & \textbf{10.46} & 17.77$\times$ \\
GRAM   & function $(32)$  & 8.91  & 31.99$\times$ \\
\midrule
MLP    & dense FP32       & 6.65  & 1.00$\times$ \\
MLP    & uniform W4 PTQ   & 6.06  & 8.00$\times$ \\
\bottomrule
\end{tabular}
\end{table}

\subsection{Sensitivity: samples and signature domain}

At $K{=}32$, accuracy is insensitive to the number of signature samples
(32 / 128 / 512 samples: 83.32 / 82.36 / 82.11\%) and to the signature domain
(grid 83.39\% vs.\ empirical activation range 82.66\%). These runs use the
base-excluded signature and so are not comparable in absolute terms to the
main pipeline; only the trend within the table is meaningful.

\subsection{Width sweep}

Compression ratios are stable across network width, and the accuracy gap to
dense closes as the network grows.

\begin{table}[h]
\centering
\caption{Branch-aware $(32,16)$ at W4 across hidden widths (MNIST, seed~42).}
\label{tab:width}
\small
\begin{tabular}{lrrrr}
\toprule
Width & 32 & 64 & 128 & 256 \\
\midrule
spline acc.\ (\%)  & 94.38 & 95.34 & 95.63 & 96.49 \\
spline comp.       & 31.28$\times$ & 31.64$\times$ & 31.82$\times$ & 31.91$\times$ \\
GRAM acc.\ (\%)    & 95.31 & 96.14 & 96.64 & 97.51 \\
GRAM comp.         & 17.46$\times$ & 17.62$\times$ & 17.70$\times$ & 17.74$\times$ \\
\bottomrule
\end{tabular}
\end{table}

\section{Convolutional KAGN Track}
\label{apx:conv}

\paragraph{Backbone and protocol.} Both CIFAR sets use the same 8-layer
convolutional KAGN (channels $[64,128,256,256,384,384,512,256]$, groups 1,
Gram degree 3, $C{=}5$ coefficients per edge): 30{,}623{,}552 parameters and
6{,}123{,}712 edges on CIFAR-10; 30{,}738{,}752 and 6{,}146{,}752 on
CIFAR-100; $116.84$ and $117.28$\,MiB FP32. Training: AdamW (lr $10^{-3}$,
weight decay $5\times 10^{-5}$), cosine schedule with 5 warmup epochs, 200
epochs, label smoothing 0.1, Cutout(8), mixup $\alpha{=}0.2$ on CIFAR-100
only, EMA 0.999, fp16, seeds 42/123/2026, splits 45k/5k/10k with selection on
the validation set. Every compressed arm starts from the same dense checkpoint
and receives the same 30-epoch fine-tune (lr $2\times 10^{-4}$); uniform PTQ
alone is inference-only by design. Storage accounting charges edge indices, codebooks, scales, prune masks,
and all normalization parameters for every method.

\begin{table}[!ht]
\centering
\caption{Conv-KAGN, three seeds, w8 codebooks, bits per edge (b/e) and top-1
accuracy. Iso-dense rows are single-seed.}
\label{tab:apx-conv}
\footnotesize
\setlength{\tabcolsep}{2.8pt}
\begin{tabular}{lccc}
\toprule
Method & b/e & CIFAR-10 & CIFAR-100 \\
\midrule
dense FP32            & 160.05 & 91.43 $\pm$ 0.21 & 62.83 $\pm$ 0.96 \\
\funccode{} fn $K{=}8$  & 3.03 & 81.18 $\pm$ 1.32 & 45.25 $\pm$ 0.85 \\
\funccode{} fn $K{=}16$ & 4.03 & 88.88 $\pm$ 0.39 & 54.19 $\pm$ 0.68 \\
\funccode{} fn $K{=}32$ & 5.03 & 90.25 $\pm$ 0.18 & 57.65 $\pm$ 0.64 \\
\funccode{} fn $K{=}64$ & 6.03 & 90.50 $\pm$ 0.41 & 59.48 $\pm$ 1.02 \\
\funccode{} fn $K{=}256$ & 8.05 & 90.89 $\pm$ 0.32 & 60.94 $\pm$ 0.79 \\
iso-dense (5\,b/e)    & 5.03 & 77.29 & 32.65 \\
iso-dense (8\,b/e)    & 8.00 & 82.38 & 37.93 \\
prune 95\% $+$ W4     & 6.04 & 90.09 $\pm$ 0.35 & 60.70 $\pm$ 1.26 \\
prune 90\% $+$ W4     & 7.04 & 90.82 $\pm$ 0.19 & 60.89 $\pm$ 1.11 \\
LSQ QAT W2            & 10.04 & 90.98 $\pm$ 0.17 & 61.34 $\pm$ 0.76 \\
LSQ QAT W4 (C100)     & 20.04 & -- & 62.24 $\pm$ 0.98 \\
uniform PTQ W2        & 10.04 & 10.00 $\pm$ 0.00 & 1.00 $\pm$ 0.00 \\
uniform PTQ W8        & 40.04 & 91.42 $\pm$ 0.24 & 62.79 $\pm$ 0.94 \\
\bottomrule
\end{tabular}
\end{table}

\paragraph{Acceptance criteria.} $K{=}256$ lands within $0.54 \pm 0.39$\,pp of
dense on CIFAR-10 and $1.89 \pm 1.24$ on CIFAR-100, passing the 3\,pp and
5\,pp allowances. At approximately 6\,b/e, \funccode{} $K{=}64$ and
95\% pruning are statistically tied on both datasets, with margins of roughly
one standard deviation. Uniform PTQ W2 collapses without fine-tuning and is
therefore reported only as a post-training quantization fragility measurement.

\paragraph{Ablations (CIFAR-100, $K{=}32$, w8, single seed).} Skipping the
first convolution and the classifier head ($0.4\%$ of edges) lifts accuracy
from $57.65 \pm 0.64$ to $60.24$ at $5.36$ against $5.03$\,b/e. The clustering
metric comparison reads: whitened function space $57.65 \pm 0.64$
(reconstruction error $0.0647$), coefficient space $59.02$ ($0.1275$), legacy
z-scored signature $54.68$ ($0.2566$); the whitening halves the error it
optimizes and does not improve accuracy. The codebook-bit sweep moves storage
by under $0.002$\,b/e from w8 to w2 while accuracy reads $58.35$ (w8),
$57.67$ (w6), $57.45$ (w4), $2.17$ (w3), $0.99$ (w2); the cliff sits between
four and three bits.

\paragraph{The w4 instability.} Four observations of the identical $K{=}32$
CIFAR-100 configuration span $9.99$\,pp at w4 ($47.46$, $51.61$, $51.83$,
$57.45$) against $1.35$\,pp at w8 ($57.67$, $57.00$, $58.28$, $58.35$), with
identical pre-quantization reconstruction error ($0.0647$) in every case. A
single w4 measurement is uninformative at this scale; all conv results are
reported at w8. Run-to-run variation at a fixed seed is $0.34$\,pp
(cudnn autotuning and non-deterministic atomics), so exact reproduction
requires deterministic backends and disabling mixed precision.

\paragraph{Disclosures.} The campaign plan specified a 4-layer
$[32,64,128,256]$ grouped backbone for CIFAR-10; it reaches $64.26\%$ against
a $91\%$ target because the shared dropout ($0.25/0.5$) is sized for the 30M
model (removing dropout recovers $+19.9$\,pp at identical parameter count,
still short at $83.92$), and it was replaced by the 8-layer net. The CIFAR-100
dense backbone runs $3.17$\,pp under its $66\%$ target ($3.3\sigma$); relative
comparisons remain valid because all arms share one checkpoint per seed and an
equal fine-tune budget. The conv branch arms are dominated by the function
codebook at every comparable budget and remain single-seed, as do the
iso-dense rows and all ablation arms. The conv study covers one polynomial
family on two backbones and two datasets.

\section{Data-Quality Anomalies}
\label{apx:anomalies}
We flag four issues openly. First, the MetaCluster spline arm fails its
fidelity gate (Section~\ref{apx:meta}) and is excluded pending a retune.
Second, the FastKAN branch $(32,16)$ configuration on CIFAR-10 reads
$47.86 \pm 0.76$ in the five-seed control family but $36.26 \pm 2.02$ in the
ten-seed benchmark, an $11.6$\,pp gap on nominally identical configurations;
until the discrepancy is resolved we treat FastKAN's branch arm as unstable
and build no claims on it in either direction. Third, the dense spline
baseline on CIFAR-100 is itself unstable across seeds ($14.21 \pm 3.15$), and
its compressed rows inherit that spread. Fourth, several legacy Fashion-MNIST
GRAM runs have dense baselines that collapse on some seeds, occasionally low
enough that a compressed model appears to beat its own dense reference; we
treat those as broken baselines rather than gains and mark the affected rows.

\section{Reproducibility}
\label{apx:repro}
The ablation archive contains 70 experiment families and 187 runs (5714 raw
result rows). Independent seed aggregation reproduces the project's own
summary tables to a maximum absolute difference of $1.4\times 10^{-14}$, and
storage columns are constant across seeds within every configuration. The
primary benchmark uses seeds 0--9; controls, low-rank, LSQ, stress and
MetaCluster studies use seeds 0--4; soft-assignment and the small-model
studies use seeds 0--2. All accuracies are computed from the released run
CSVs, and every storage figure comes from the bit-exact packed export.

\end{document}